\documentclass[sigconf]{acmart}

\usepackage{hyperref}
\usepackage{url}
\usepackage{subfigure}
\usepackage{hyperref}
\usepackage{xcolor,soul,colortbl}
\usepackage{booktabs} 
\usepackage{enumitem}
\usepackage{graphicx}
\usepackage{multirow}
\usepackage{amsfonts}
\usepackage{array}
\usepackage{amsthm}
\theoremstyle{definition}

\usepackage{wrapfig}
\usepackage{lipsum}
\usepackage{bbding}

\usepackage{algorithm}
\usepackage{algorithmic}
\usepackage[algo2e,ruled,linesnumbered]{algorithm2e}

\newcommand{\revise}[1]{\textcolor{black}{{#1}}}
\AtBeginDocument{%
  }

\copyrightyear{2024} 
\acmYear{2024} 
\setcopyright{rightsretained} 
\acmConference[KDD '24]{Proceedings of the 30th ACM SIGKDD Conference on
 Knowledge Discovery and Data Mining}{August 25--29, 2024}{Barcelona, Spain}
\acmBooktitle{Proceedings of the 30th ACM SIGKDD Conference on Knowledge
 Discovery and Data Mining (KDD '24), August 25--29, 2024, Barcelona,
 Spain}\acmDOI{10.1145/3637528.3671899}
\acmISBN{979-8-4007-0490-1/24/08}

\begin{document}

\title{FLea: Addressing Data Scarcity and Label Skew in Federated Learning via Privacy-preserving Feature Augmentation}

 
\author{Tong Xia}
\email{tx229@cam.ac.uk}
\affiliation{%
  \institution{University of Cambridge}
 \country{Cambridge, UK}
}

 \author{Abhirup Ghosh}
\affiliation{%
 \institution{University of Birmingham} 
 \institution{University of Cambridge}
 \country{UK}}

 \author{Xinchi Qiu}
\affiliation{%
 \institution{University of Cambridge}
 \country{Cambridge, UK}}

\author{Cecilia Mascolo}
\affiliation{%
 \institution{University of Cambridge}
 \country{Cambridge, UK}}

\renewcommand{\shortauthors}{Tong Xia et al.}

\begin{abstract}
Federated Learning (FL) enables model development by leveraging data distributed across numerous edge devices without transferring local data to a central server. However, existing FL methods still face challenges when dealing with scarce and label-skewed data across devices, resulting in local model overfitting and drift, consequently hindering the performance of the global model.
In response to these challenges, we propose a pioneering framework called \textit{FLea}, incorporating the following key components:
\textit{i)} A global feature buffer that stores activation-target pairs shared from multiple clients to support local training. This design mitigates local model drift caused by the absence of certain classes;
\textit{ii)} A feature augmentation approach based on local and global activation mix-ups for local training. This strategy enlarges the training samples, thereby reducing the risk of local overfitting;
\textit{iii)} An obfuscation method to minimize the correlation between intermediate activations and the source data, enhancing the privacy of shared features.
To verify the superiority of \textit{FLea}, we conduct extensive experiments using a wide range of data modalities, simulating different levels of local data scarcity and label skew. The results demonstrate that \textit{FLea} consistently outperforms state-of-the-art FL counterparts (among 13 of the experimented 18 settings, the improvement is over $5\%$) while concurrently mitigating the privacy vulnerabilities associated with shared features. Code is available at \url{https://github.com/XTxiatong/FLea.git}.
\end{abstract}



\maketitle

\section{Introduction}

Presently, there are billions of interconnected edge devices, including smartphones, tablets, and wearables, generating a continuous stream of data such as photos, videos, and audio~\cite{lim2020federated}. Such data presents numerous opportunities for meaningful research and applications. However, the conventional approach of aggregating this data in a central server is no longer sustainable, as the data can be sensitive to share and the communication cost associated with transferring such vast amounts of information can be prohibitive. Thanks to the advent of Federated Learning (FL), developing models with data remaining at edge devices becomes feasible~\cite{liu2021keep}.

\begin{figure} [t]  
  \begin{center}
 \includegraphics[width=0.45\textwidth]{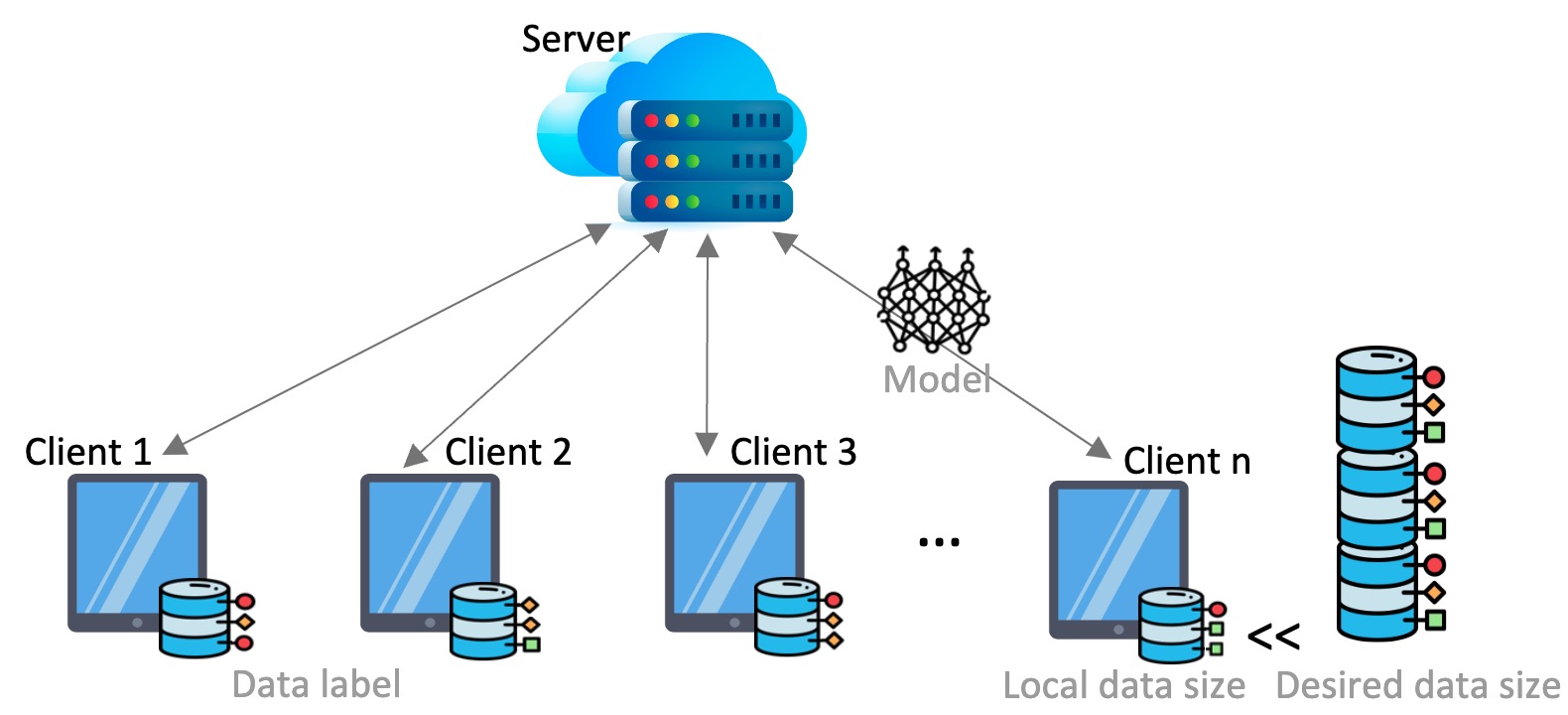}
  \end{center}
  \vspace{-10pt}
  \caption{ Edge devices as clients in federated learning, where local data exhibits label skew (presented by different markers) and scarcity (usually very small in size).}
    \label{fig:FL}
    \vspace{-15pt}
\end{figure}

As illustrated in Figure~\ref{fig:FL}, in FL,  edge devices, usually referred to as \textit{clients}, train \textit{local models} using their local data. These trained models are then transmitted to the FL \textit{server} and aggregated into a \textit{global model} for real-world applications~\cite{mcmahan2017communication}. 
However, the characteristics of the data across edge devices can present significant challenges to FL, mainly because of:

\textit{i) Data heterogeneity, specifically label-skewness.} 
As shown in Figure~\ref{fig:FL}, edge devices may only possess a subset of the global categories in their acquired datasets. For example, in terms of images, one mobile user may take photos of cats and indoor decorations, while another user may have images of dogs and outdoor views. Such label-skewness can result in local model drift: local models are biased toward the local distribution and struggle to generalize to the global distribution, leading to a sub-optimal global model~\citep{li2020federated,karimireddy2020scaffold,luo2021no}.

\textit{ii) Local data scarcity.} Datasets collected by edge devices are also limited in size, primarily due to limited data acquisition scenarios and the difficulty in annotations. As shown in Figure~\ref{fig:FL}, local data could be much smaller than the desired data to optimize the local models. In the prominent FL strategy \textit{FedAvg}~\citep{mcmahan2017communication}, lower aggregation weights are assigned to local models trained on smaller datasets, indicating their relatively weaker performance and lesser contribution to the global model. However, in a situation where all clients possess small-sized datasets, local models can be overfitted to the training samples and thus struggle to generalize to unseen testing data, even if from the same distribution. Consequently, aggregating these models does not improve the global model's generalization ability. This, as a result, slows down convergence and negatively impacts the performance of the global model (a detailed analysis of the effect of data scarcity is presented in Sec.~\ref{Sec:scarcity}).

To fully harness the potential of FL for real-world edge applications, addressing the two challenges mentioned above is pivotal. Despite numerous proposed approaches aimed at enhancing FL in the presence of label skew, we find that the feasibility of these methods, when confronted with data scarcity, remains an under-explored area. Methods that involve devising new learning objectives~\cite{li2021fedrs} or aggregation strategies~\cite{yurochkin2019bayesian} falter when local data is extremely scarce, as they struggle to find a balance between local optimization and preserving global knowledge. On the other hand, certain approaches leverage globally shareable information~\cite{zhao2018federated}, such as a proportion of raw data, which may be effective when the data across edge devices are scarce and label-skewed. However, these strategies often come at the cost of sacrificing privacy. 
In light of this, we revisit the purpose of information sharing and pose a fundamental question: \textbf{\textit{Can we generate and share some privacy-preserving information as a robust proxy for global data distribution to aid local learning?}} This, in turn, could alleviate issues of local overfitting and model drift caused by simultaneous challenges of data scarcity and label skew.

The shared information, referred to as \textit{features} in this study, should satisfy the following critical criteria:  (1) To alleviate local drift caused by label skew, it should cover all categories. (2) To prevent local overfitting due to data scarcity, it needs to be useful in extending the local training data. (3) To mitigate privacy risks, it requires to contain minimal information from the raw data. With these considerations in mind, we introduce a novel method \textit{FLea} (\underline{FL} with f\underline{ea}ture sharing). In \textit{FLea}, the shared features are activations from an intermediate layer of the model and the associated labels. Specifically, we maintain a global feature buffer that includes features from multiple clients, ensuring coverage of all categories. The shared features are integrated with local data through a mix-up augmentation~\cite{zhang2018mix} to extend the training samples in the representation space. Moreover, these activations undergo a level of ``obfuscation'' by reducing their distance correlation with the source data~\cite{vepakomma2020nopeek}, while maintaining their classification characteristics through a customized loss function during training.

Overall, this paper has the following contributions:
\begin{itemize}[topsep=-2pt, itemsep=0pt,  leftmargin=20pt]
    \item The first in-depth study on a common but under-explored scenario in FL, where all the clients possess extremely scarce and label-skewed data. We find that model overfitting caused by data scarcity is under-looked by existing methods. 
    \item A novel framework \textit{FLea} to address both the data scarcity and label skew challenges at the same time in FL. The key idea of \textit{FLea} is to share privacy-preserving features to augment local data for model training. 
    \item Extensive experiments with different levels of label skew and data scarcity show that \textit{FLea} consistently outperforms the state-of-the-art baselines. We also empirically demonstrate that \textit{FLea} mitigates privacy risk measured by information leakage from the shared features.
\end{itemize}

\section{Background}
\subsection{Fundamentals of FL}\label{FedAvg}
This study focuses on learning a global model from a set of collaborating clients, $K$, with each client $k$ containing a small and label-skewed local dataset $\mathcal{D}_k$. The FL system works in synchronous rounds. At the start of each round $t$, the FL server broadcasts the current global model parameters $\theta^{(t)}$ to the randomly selected subset of the clients $\mathcal{K}^{(t)} \subseteq K$.
Each client $k \in \mathcal{K}^{(t)}$ takes a few optimization steps (e.g., using stochastic gradient descend) starting from $\theta^{(t)}$, resulting in an updated local model $\theta^{(t)}_k$. The local optimization aims to minimize the loss function $\mathcal{L}$ on local data $\mathcal{D}_k$, i.e., 
$\theta_k = \arg \min_{\theta} \mathcal{L}(\theta, \mathcal{D}_k| \theta^{(t)})$ ($\mathcal{L}$ is the learning objective which is usually a cross-entropy loss for classification).
Each round $t$ ends with model aggregation to derive the new global model $\theta^{(t+1)}$. 
The most basic and popular aggregation method, \emph{FedAvg}~\citep{mcmahan2017communication} averages the model parameters weighted by the fraction of local data size in the clients,
 \begin{equation}\label{eq:fedavg-aggr}
 \vspace{-3pt}
    \theta^{(t+1)} = \sum_{k \in \mathcal{K}^{(t)}} \frac{|\mathcal{D}_k|} { \sum_{k\in \mathcal{K}^{(t)}} |\mathcal{D}_k|} \theta_k.
\end{equation}

\subsection{Addressing label skew in FL}\label{Sec:related_skew}

To mitigate the client drift caused by label skew, many methods have been proposed to \textit{improve the learning objective} $\mathcal{L}$. 
In addition to classification loss, \textit{FedProx}~\citep{li2020federated} also regulates the discrepancy between the local and global model parameters. \textit{MOON}~\citep{li2021model} leverages constructive learning to maximize the distance between low-dimensional features and other classes, thereby improving feature learning.
To compensate for missing categories in the local data, \textit{FedDecorr}~\citep{shi2022towards} introduces regularization of the dimensional collapse in FL models, while \textit{FedNTD}~\citep{lee2022preservation} penalizes changes in the non-ground-truth class \textit{logit} distribution predicted by global and local models.  
\textit{FedLC}~\citep{zhang2022federated} directly re-scales the logits to derive a calibrated loss. Similarly, \textit{FedRS} restricts the \textit{Softmax} to limit the update of missing classes. 

\textit{Data augmentations} have been explored for label skewed FL.
\citep{zhao2018federated} shows that sharing a small proportion of local data globally, alongside the model parameters, can significantly enhance \textit{FedAvg} (in later sections, we name this method \textit{FedData}). 
Nevertheless, collecting private data would compromise the privacy-preservation benefits of FL. Therefore, other global proxies that are less privacy-sensitive than raw data are explored. \textit{FedMix}~\citep{yoon2020fedmix} and \textit{FedBR}~\citep{guo2023fedbr} average data over mini-batches and share this aggregated data globally, while \textit{CCVR}~\citep{luo2021no} shares low-dimensional features with the server to calibrate the global model on the server side. 
 These low-dimensional features are also known as class prototypes, which are explored to mitigate local classifier bias~\citep{tan2022fedproto}.
\textit{FedGen}~\citep{liu2022overcoming},  VHL~\citep{tang2022virtual}, and \textit{FedFed}~\cite{yang2023fedfed} generate data samples through an additional generative model to aid local learning. Although no raw information is shared by those two approaches, their performance highly depends on the quantity and quality of synthetic data, usually yielding marginal gains over \textit{FedAvg}.

\subsection{Data scarcity in FL}\label{Sec:scarcity}
Through the literature review, we have found few studies focusing on the data scarcity problem in FL. As first observed by~\citep{li2022federated} (cf. Finding 7), when developing the model with a fixed total number of training samples, the accuracy of \textit{FedAvg} and \textit{FedProx} decreases as the number of clients increases, leading to a reduction in local data size per client. Despite the proliferation of FL methods, including those mentioned earlier (cf. Sec.~\ref{Sec:related_skew}), most are evaluated on large local datasets, each containing thousands of samples. For example, the commonly used benchmark CIFAR10 contains $50,000$ training samples, which are usually distributed among $10$~to~$100$ clients, resulting in an average local data size of $500$~to~$5,000$ \cite{li2021fedrs, guo2023fedbr, lee2022preservation}. In contrast, in a scarce setting, the average local data size could be even much smaller, such as $50\sim 100$. This leaves the effectiveness of existing FL methods in handling data scarcity unclear. While some studies have presented results in a scarce setting~\citep{charles2021large,zhu2023confidence}, they fail to justify whether the performance gain is from alleviating bias or overfitting caused by data scarcity, thus lacking in providing a deep understanding. 

\textit{FedScale} introduced the first benchmark featuring thousands of clients with limited training data~\citep{lai2022fedscale}. However, \textit{FedScale} primarily focuses on system efficiency and offers limited insight into algorithm effectiveness. On the contrary, in the following, we conduct an empirical study and provide insightful observations on how data scarcity can affect the performance of FL.

 \subsubsection{Performance decline caused by data scarcity}\label{Sec:2.3.2}

To directly observe how data scarcity affects FL, we conducted experiments comparing several state-of-the-art FL algorithms at different data scarcity levels. Specifically, we compared \textit{FedAvg} with the best loss-based methods \textit{FedDecorr} and \textit{FedNTD}, as well as the best data augmentation-based methods \textit{FedMix} and \textit{FedData}.  To avoid being affected by label skew, we split the data in an IID (Independent and Identically Distributed) manner. We distributed the training set of CIFAR10 to $10$, $100$, and $500$ clients, leading to a local data size $|\mathcal{D}_k|$ of $5000$ (not scarce), $500$ (mildly scarce), and $100$ (scarce), respectively, ensuring a uniform representation of all 10 classes. More experimental details are presented in Appendix~\ref{Sec:Apen_A}.

\begin{figure}
    \centering 
  \vspace{-5pt}  
\includegraphics[width=0.49\textwidth]{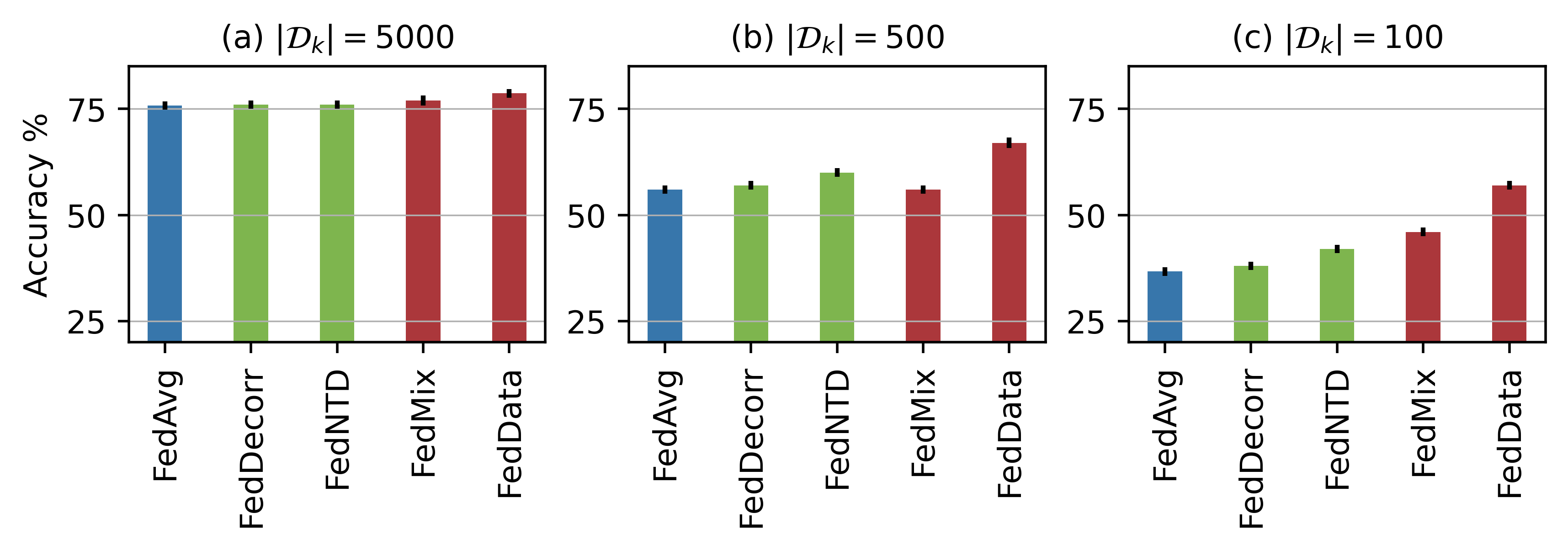}
\vspace{-20pt}
    \caption{Performance of FL methods with increasing data scarcity levels (A smaller $|\mathcal{D}_k|$  indicates a heavier scarcity).}\label{fig:exist}
    \vspace{-16pt}
\end{figure}

The results are summarized in Figure~\ref{fig:exist}, from which we draw the following observations:
\textbf{1)\textit{FedAvg} degrades remarkably as data scarcity becomes more severe}: Its accuracy, which starts at $75\%$ in (a) with sufficient data, decreases to $56\%$ in (b) when $|\mathcal{D}_k|$ is reduced to $500$, and further drops to $37\%$ in (c) when $|\mathcal{D}_k|$ is reduced to $100$.
\textbf{2) The compared loss-based methods also decline as the data become scarce}: From (a) to (c), although \textit{FedDeccor} and \textit{FedNTD} outperform \textit{FedAvg}, the gain is marginal, and all their accuracy drops significantly.
\textbf{3) The compared data augmentation-based methods outperform other methods}: With internal data exchange, \textit{FedMix} and \textit{FedData} effectively improve local models, leading to remarkable performance gains over \textit{FedAvg}. When data scarcity is significant ($|\mathcal{D}_k|=100$), they notably outperform loss-based FL methods, as shown in (c).
 
 \subsubsection{Understanding the effect of data scarcity}\label{Sec:DB}
From the above observation, we hypothesize that \textbf{data scarcity can lead to local model overfitting}, and aggregating such models cannot continually improve the global model. In light of this, the data augmentation-based methods can prevent the performance decline as more training samples are available to alleviate the local overfitting.  To verify this, we carried out controlled experiments on CIFAR10 to analyze the model update in one communication round with different amounts of training samples. 

 \begin{figure}
    \includegraphics[width=0.48\textwidth]{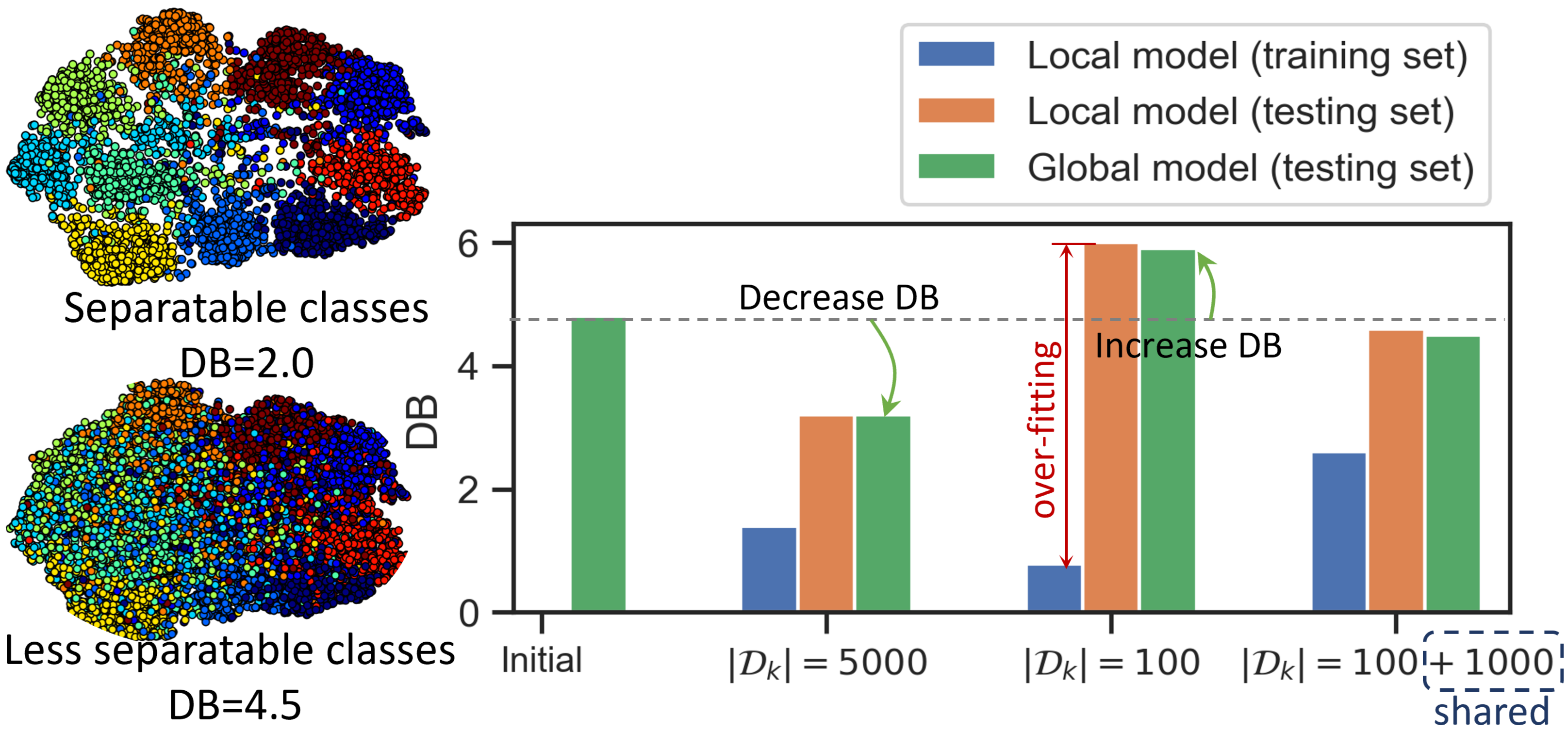}
  \vspace{-15pt}
  \caption{T-SNE for low-dimension features where the color distinguishes classes and the class separation measurement DB under different numbers of training samples.}
    \label{fig:dbs}
    \vspace{-12pt}
\end{figure}

\begin{figure}[t]
    \centering 
\includegraphics[width=0.49\textwidth]{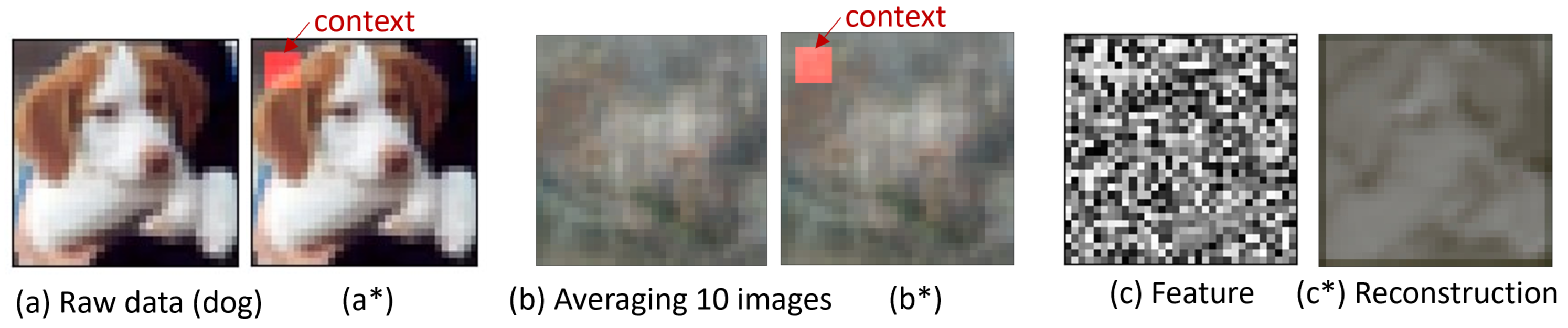}
\vspace{-15pt}
    \caption{Data augmentations. From (a) to (c), the privacy vulnerability is reduced.  (b) is the average of a batch of samples like (a), but if the local data contains individual context information (e.g., (a*)), averaging over those samples cannot protect such information (e.g., (b*)).  (c) shows a feature of (a*) and (c*) shows its reconstruction.}\label{fig:datashared}
    \vspace{-15pt}
\end{figure}

Specifically, we examined a communication round where local training begins with a global model possessing an accuracy of $40\%$. Following the IID setting outlined in Sec.~\ref{Sec:2.3.2}, we compare the performance of local models trained with $|\mathcal{D}_k|=5000$ and $|\mathcal{D}_k|=100|$ samples. To assess the models, we scrutinized the quality of learned activations across different classes.
For visualization, we adopted a commonly used tool that maps activations from the penultimate layer of the model into a two-dimensional space~\citep{van2008visualizing,guo2023fedbr}. Quantitatively, we employed the clustering score, specifically the Davies-Bouldin Score (DB), to measure the separation of activations among classes. A detailed formulation can be found in Appendix~\ref{Sec:Apen_A}.
As illustrated in Figure~\ref{fig:dbs} (left), a smaller DB indicates less overlap among activations. The DBs before and after local training are summarized in the bar chart presented in Figure~\ref{fig:dbs}.

For local models, after training, the DB was reduced from 4.8 to 3.1 when $|\mathcal{D}_k|=5000$ and to 0.8 when $|\mathcal{D}_k|=100$ on the training set. However, \textbf{a notable gap between DB values on the training set and testing set can be observed, suggesting the occurrence of overfitting.} As the number of samples decreases ($|\mathcal{D}_k|=100$), the gap widens, indicating severe overfitting. Consequently, after aggregation, the performance of the global model varies, and training with only $|\mathcal{D}_k|=100$ fails to enhance the global model, as evidenced by an increased DB. These results uncover the negative impact of data scarcity on the generalization of local models, resulting in overlapped features and ultimately leading to an under-performing global model.

Furthermore, in the fourth group ($|\mathcal{D}_k|=100+1000$) where local data contain $100$ samples but a globally shared set with $1000$ samples is also used for training,  a smaller gap between the training and testing sets is observed. This indicates that with more shared data to aid local training, the generalization of the local model improves. Consequently, the global model's performance is enhanced, as evidenced by a decreased DB. This observation supports the feasibility of data augmentation-based methods for addressing data scarcity in FL.

\subsection{Privacy-preserving feature sharing}\label{Sec:2.4}
The above analysis highlights the challenge of overfitting caused by data scarcity, and further suggests that globally sharing certain information can help mitigate this problem. In Sec.~\ref{Sec:related_skew}, several data augmentation-based FL methods are introduced; however, it is crucial to note that these methods may introduce privacy vulnerabilities.

As illustrated in Figure~\ref{fig:datashared}, \textit{FedData} globally shares raw data and labels, while \textit{FedMix} shares aggregated data and labels globally. Although the averaging of samples in \textit{FedMix} hinders data reconstruction, it remains privacy-vulnerable, as it releases contextual information. To illustrate, consider a scenario where a client's phone camera has a sensor problem, resulting in a spot in each photo (see Figure~\ref{fig:datashared}(a*)). Alternatively, imagine a client residing in a bustling neighborhood, leading to a constant background score in all audio clips. Averaging over a batch of samples fails to protect such specific context information, as depicted in Figure~\ref{fig:datashared}(b*). 
Unlike these existing works, to improve the trade-offs between performance and privacy protection, we propose to share the activations from the intermediary layers (see Figure~\ref{fig:datashared}(c))). To enhance the privacy measured by the information leakage from those activations, we employ an obfuscation approach to reduce the distance correlation between activations and the source data. One example of data reconstructed from the activation of is shown in Figure~\ref{fig:datashared}(c*), where sensitivity information like color and context are not recovered.

\section{FLea}\label{Sec:method}
\subsection{Overview} 
Building upon the insights gained from the preceding discussion and analysis, we now formally introduce our method, \textit{FLea}. It aims to address the challenges of data scarcity and label skew while minimizing privacy risks associated with shared information.

At an abstract level, \emph{FLea} maintains a feature buffer containing activation-target pairs from multiple clients. This shared buffer enables clients to have more training samples covering all classes for local training. To safeguard the privacy of shared features, we obfuscate activations by minimizing the correlation between activations and the source data when extracting these activations. 

\begin{figure}
    \centering
    \vspace{-30pt}
\includegraphics[width=0.5\textwidth]{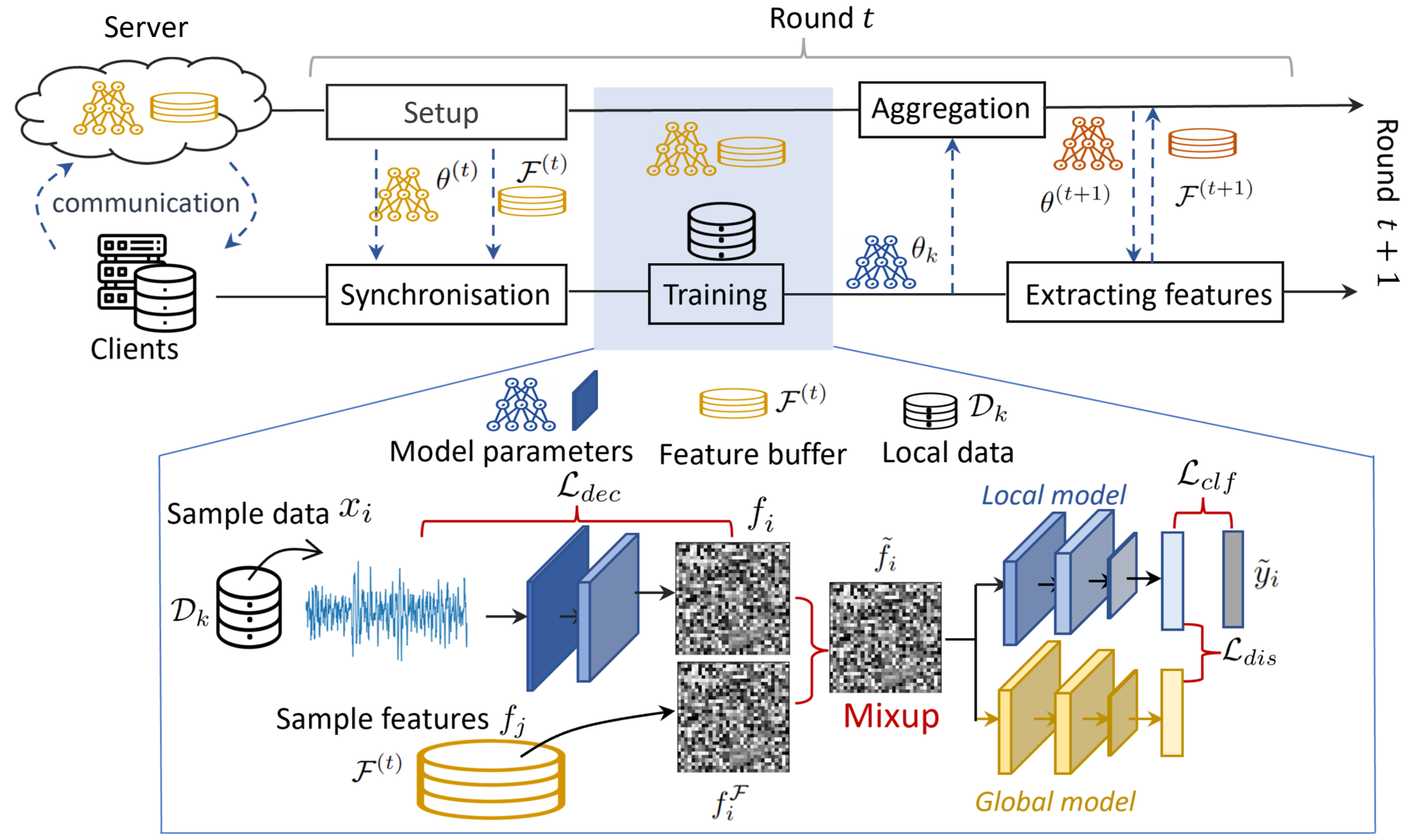}
\vspace{-15pt}
    \caption{Overview of \textit{FLea} for $t$-th communication round.}\label{fig:framw} 
\vspace{-15pt}
\end{figure}

Examining the training process, \emph{FLea} operates iteratively, akin to \textit{FedAvg}. Initially, the global model is randomly initialized, and the buffer is empty. Then, for each round $t$, as illustrated in Figure~\ref{fig:framw}, \textit{FLea} starts with synchronizing the global model parameters $\theta^{(t)}$ and feature buffer $\mathcal{F}^{(t)}$ to the selected clients $\mathcal{K}^{(t)}$. Once local training using $\mathcal{D}_k$ and $\mathcal{F}^{(t)}$ completes (the first round only uses $\mathcal{D}_k$ since the feature buffer is empty), those clients send the updated model parameters $\theta_k$ to the server, to be aggregated into a new global model parameterized by $\theta^{(t+1)}$.  Followed by that, \textit{FLea} needs another step to update the global feature buffer to $\mathcal{F}^{(t+1)}$. A detailed training procedure is summarized in Algorithm~\ref{al:train}. We elaborate on the main components of the procedure in the following sections.

\subsection{Formulation of feature buffer}
The shared feature buffer contains activation-target pairs from multiple clients. An activation is the intermediate-layer output of a model.
Let's consider the model parameters $\theta$ to be divided into two parts at layer $l$: $\theta[:l]$ and $\theta[l:]$. For client $k$, the activation extracted from data $x_i\in \mathcal{D}_k$ is $\theta[:l](x_i) = f^{\mathcal{F}}_i$. The feature buffer from this client is the set of pairs including activations and their labels, termed as $(f^{\mathcal{F}}_i, y^{\mathcal{F}}_i)$. Each client randomly selects a $\alpha$ fraction of its local data to create its feature buffer to share with others. The server gathers those local feature buffers and merges them into the global one $\mathcal{F}$. Although data from a single client is skewed, aggregating features from multiple clients can cover many classes and thus alleviate the label-skew problem. 
Note that a client only extracts and contributes to the global feature buffer at the round when it participates in training and the global buffer resets at every round. This reduces the exposure and mitigates the privacy risk.

\subsection{Client local training}
Suppose client $k$ is selected in round $t$, i.e., $k \in \mathcal{K}^{(t)}$. As shown in Figure~\ref{fig:framw}, $k$ receives  the global model $\theta_k= \theta^{(t)}$ and the feature buffer $\mathcal{F}^{(t)}$.  The local data $\mathcal{D}_k$ and the feature buffer $\mathcal{F}^{(t)}$ are divided into equal-sized batches for model optimization, termed by $\mathcal{B}=\{(x_i,y_i) \in \mathcal{D}_k\}$ and $\mathcal{B}^{\mathcal{F}}=\{(f^{\mathcal{F}}_i,y^{\mathcal{F}}_i)  \in \mathcal{F}^{(t)}\}$, respectively ($|\mathcal{B}| = |\mathcal{B}^{\mathcal{F}}|$).  The traditional method will feed $\mathcal{B}$ into the model directly to optimize the model, but to address the data scarcity and label skew problem, we propose to augment the input in the representation space. 

\subsubsection{Feature augmentation}\label{sec:aug}
We feed $\mathcal{B}$ into the first $l$ layers of the model, extracting the intermediate output for each data point, and then we generate the augmented inputs $\mathcal{\tilde{B}}$.
Specifically, for each data sample $(x_i, y_i)$ in $\mathcal{B}$, we extract its feature $(f_i=\theta_k[:l](x_i), y_i)$ and mix the feature with a sample $(f^{\mathcal{F}}_i,y^{\mathcal{F}}_i)$ from the feature buffer batch $\mathcal{B}^{\mathcal{F}}$. The generated augmentation $(\tilde{f}_i, \tilde{y}_i)$ for $i$-th sample is formulated as (the new batch is termed as $\mathcal{\tilde{B}}$),
\begin{equation}\label{mixup}
\begin{cases}
  \ \tilde{f}_i = \beta_i f_i + (1-\beta_i) f^{\mathcal{F}}_i, \\
  \ \tilde{y}_i = \beta_i y_i + (1-\beta_i) y^{\mathcal{F}}_i, \\
\end{cases}
\end{equation}
where $\beta_i$ is the weight balancing the strength of interpolation between the local and global features.
Inspired by the data augmentation method in the centralized setting~\citep{zhang2018mix}, we sample the weight $\beta_i$  for each data point from a symmetrical Beta distribution~\citep{gupta2004handbook}: $\beta_i \sim Beta(a,a)$.   $\beta_i \in [0,1]$ controls the strength of interpolation: A smaller $\beta_i$ makes the generated sample closer to the local feature while a larger one pushes that to the global feature.  $a$ is a hyper-parameter controlling the shape of the Beta distribution.

Since the data batch $\mathcal{B}$ and the feature batch $\mathcal{B}^{\mathcal{F}}$ are randomly shuffled, and $\beta_i$ is sampled from the Beta distribution for each pair, the augmentation encounters an unlimited combination of local and shared features, thereby significantly extending the classes and the number of samples for model training.

\subsubsection{Local training objective}
As illustrated in Figure~\ref{fig:framw}, the augmented features in $\mathcal{\tilde{B}}$ are then inputted into the model starting from layer $l$. For $\tilde{f}_i$, let the output logit be denoted by $z^l_i$: $z^l_i = \theta_{k}[l:](\tilde{f}_i)$, and the probability for class $c$ is $p^l_i[c] = \frac{\exp(z^l_i[c])}{\sum_c \exp(z^l_i[c])}$. Given this, we devise a loss function comprising multiple components to optimize the local model.
The first term aims to minimize the classification error, which is formulated as,
\begin{equation}
  \mathcal{L}_{clf}(\mathcal{B},\mathcal{B}^{\mathcal{F}}) = \frac{1}{|\mathcal{B}|}\sum_i \sum_c - \tilde{y}_i[c] \log p^l_i[c].
\end{equation}
It is equivalent to the cross-entropy loss.
The second loss term is used to distill the knowledge from the previous global model to prevent local drift, and is derived by the KL-divergence between the global probabilities and local probabilities as~\citep{hinton2015distilling},
\begin{equation}
  \mathcal{L}_{dis}(\mathcal{B},\mathcal{B}^{\mathcal{F}}) = \frac{1}{|\mathcal{B}|}\sum_i \sum_c  - p^l_i[c] \log \frac{p^g_i[c]}{p^l_i[c]}, 
\end{equation}
where for $\tilde{f}_i$ the global logit is $z^g_i = \theta^{(t)}[l:](\tilde{f}_i)$ and the global probability is $p^g_i[c] = \frac{exp(z^g_i[c])}{\sum_c exp(z^g_i[c])}$.  Mitigating this loss encourages the local model to make similar predictions to the global model.

Besides, we aim to reduce the information leakage of the features before they are shared with other clients. As such, we learn the first $l$ layers while reducing the distance correlation
 between the activations and the source data. Thus, the third term is formulated by distance correlation~\citep{vepakomma2018supervised,vepakomma2020nopeek}, 
 \begin{equation}\label{Eq:Dec}
\mathcal{L}_{dec}(\mathcal{B}) =  \frac{\nu^2(x,f)}{\sqrt{\nu^2(x,f)\nu^2(f,f)}},
\end{equation}
where $\nu^2(,)$ denotes the squared distance. Specifically, $\nu^2(x,f) = \frac{1}{|\mathcal{B}|^2}\sum_{i,k}^{|\mathcal{B}|} \hat{E_x}[i,k]\hat{E_f}[i,k]$.  $E_x$ is the Euclidean distance matrix for $x\in \mathcal{B}$, i.e., $E_x[i,k]=||x_i - x_k||^2$. Similarly, $E_f[i,k]=||f_i - f_k||^2$. They are then double-centered to $\hat{E_X}$ and $\hat{E_F}$, by making their row and columns sum zeros (cf. Eq.~(\ref{eq:double_x}) and Eq.~(\ref{eq:double_f}) in Appendix~\ref{Sec:Apen_B})). 
After normalization, the distance correlation $c=\mathcal{L}_{dec}$ has the following properties: (1) $c$ satisfies the relation $0 \leq c \leq1$, and a smaller $c$ suggests less mutual information between $x$ and $f$; (2) $c=1$ when $f$ is a linear transformation from $x$~\cite{szekely2007measuring}. In our case, the model contains $l$ non-linear layers and thus $c<1$; (3) $c=0$ when $x$ and $f$ are independent. In other words, $c=0$ if $f$ becomes random noise and this produces perfect privacy but $f$ is useless for classification. Overall,  we aim to reduce $c$ to project feature privacy.  
Our optimization function therefore is, 
\begin{equation}\label{Eq:loss}
 \mathcal{L} =  \mathcal{L}_{clf}(\mathcal{B},\mathcal{B}^{\mathcal{F}}) + \lambda_1  \mathcal{L}_{dis}(\mathcal{B},\mathcal{B}^{\mathcal{F}}) + \lambda_2 \mathcal{L}_{dec}(\mathcal{B}),
\end{equation} 
where $\lambda_1$ and $\lambda_2$ are the weights to trade-off classification utility and privacy preservation.
The local update is then achieved by $\theta_k \leftarrow \theta_k- \eta \frac{\partial \mathcal{L}}{ \partial \theta_k }$,  where $\eta$ controls the learning rate.

\begin{algorithm2e}[t]
    \DontPrintSemicolon
    \caption{FLea}
    \label{al:train}
    \SetKwInOut{Input}{Input}
    \SetKwInOut{Output}{Output}
    \Input{Local dataset $\mathcal{D}_k$, randomly initialized model $\theta^{(1)}$. }
    \Output{Global model $\theta^{(T)}$.}
    \For {each round $t$ = 1,2,...,T }  
    { Server samples clients $\mathcal{K}^{(t)}$ and broadcasts $\theta_k \leftarrow   \theta^{(t)} $ \\
      Server broadcasts  $\mathcal{F}^{(t)}$ to $\mathcal{K}^{(t)}$   \ \ \ // Skip if $t=1$.  \\
      \For {each client $k \in \mathcal{K}^{(t)}$ in parallel }  
     { \For {local epoch $e= 1,2,..,E$}
       { \For {local batch $b = 1,2,...$}
       { Sample one data batch $\mathcal{B}$, one feature batch $\mathcal{B}^{\mathcal{F}}$, and $ \beta_i \sim Beta(a,a)$ \\ 
       Generate augmentation and update local model: $\theta_k \leftarrow \theta_k- \eta \nabla \mathcal{L} (\theta_k)$ according to Eq.~(\ref{Eq:loss}) \ \ // Skip augmentation if $t=1$\\
    }}
    Client $k$ sends $\theta_k$ to server \\
    } 
    Server aggregates $\theta_k$ to $\theta^{(t+1)}$ refer to Eq.~(\ref{eq:fedavg-aggr})

    \For {each client $k \in \mathcal{K}^{(t)}$ in parallel } 
     {  
        Client $k$ receives the new model $\theta^{(t+1)}$ \\  
       Client $k$ extracts (without gradients) and sends $\mathcal{F}_k^{(t+1)}$ to server
     }  

     Server aggregates $\mathcal{F}_k^{(t+1)}$ and update the global feature buffer to $\mathcal{F}^{(t+1)}$
     }
     \vspace{-3pt}
\end{algorithm2e}

\subsection{Model aggregation and buffer updating}\label{Sec:3.4}
Once the local training completes, the updated model parameters will be sent to the server for aggregation (lines 13 in Algorithm~\ref{al:train}). \textit{FLea} utilizes the same aggregation strategy as \textit{FedAvg} (Eq.(\ref{eq:fedavg-aggr})). The new model will be synchronized to the clients, and the features will be extracted and sent to the server to replace the old ones, updating the feature buffer to  $\mathcal{F}{(t+1)}$ (lines 14-18 in  Algorithm~\ref{al:train}). The iterations continue (restart from line 2) until the global model converges.

\section{Evaluation}
In this section, we conduct extensive experiments using three datasets to answer the following research questions:
\begin{itemize}[topsep=-2pt, itemsep=0pt,  leftmargin=10pt]
    \item \textit{\textbf{RQ1}:}  How does \textit{FLea} perform compared to the state-of-the-art FL baselines for various levels of label skew and data scarcity?
    \item \textit{\textbf{RQ2}:}  How do the main components and key hyper-parameters affect \textit{FLea}'s performance? 
    \item \textit{\textbf{RQ3}:} Can \textit{FLea}'s strategy mitigate the privacy vulnerability associated with feature sharing?
\end{itemize}

\subsection{Experimental setup}\label{Sec: setup}
 
\subsubsection{Datasets.}  We evaluate \textit{FLea} on three data modalities. \textbf{Images}: CIFAR$10$~\cite{cifar10} is a commonly used FL benchmark containing 10 classes of objectives. To classify those images, we use MobileNet\_V2~\citep{sandler2018mobilenetv2} with $18$ blocks consisting of multiple convolutional and pooling layers. \textbf{Audio}: UrbanSound$8$K~\cite{salamon2014dataset} is an audio classification bookmark also containing 10 categories of sounds collected in urban environments. Those audio samples are first transformed into spectrograms and fed into a 4-layer CNN model called \textit{AudioNet} for classification~\cite{audionet}.  \textbf{Sensory data}:  UCI-HAR~\cite{misc_human_activity_recognition_using_smartphones_240} is a public human activity recognition database collected by a waist-mounted smartphone with an embedded accelerometer. Six activities including walking, walking upstairs, walking downstairs, sitting, standing, and lying were recorded. We employ \textit{HARNet} which comprises 4 convolutional layers to recognize those activitie~\cite{teng2020layer}. 
A summarily of those datasets can be found in Appendix~\ref{Sec:Apen_C} Table~\ref{tab:data}, and the details of the models are presented in Appendix~\ref{Sec:Apen_C2}.

\subsubsection{FL setup and baselines.} 
To simulate label skew, we consider quantity-based skew (Qua($q$) with $q$ being the number of presented classes) and distribution-based skew (Dir($\mu$) with $\mu$ controlling the class skewness)~\citep{zhang2022federated}. To mimic different levels of data scarcity, we distribute the training data to $|K|$ clients, where $|K|$ is determined by the average local data size $\bar{|\mathcal{D}_k|}$, which is set to as small as $100$ and $50$ for experiments. 
A visualization of the data distribution can be found in Appendix~\ref{Sec:Apen_C1}. 

We compare~\emph{FLea} against~\emph{FedAvg}, and then \emph{loss-based} methods: $i)$~\emph{FedProx}~\citep{li2020federated}, 
$ii)$~\emph{FedDecorr}~\citep{shi2022towards},
$iii)$~\emph{FedLC}~\citep{zhang2022federated}, and 
$iv)$~\emph{FedNTD}~\citep{lee2022preservation},
as well as
\emph{data augmentation-based} methods: 
$i)$~\emph{FedBR}~\citep{guo2023fedbr},
$ii)$~\emph{CCVR}~\citep{luo2021no},
$iii)$~\emph{FedGen}~\citep{liu2022overcoming},
and $iv)$~\emph{FedMix}~\citep{yoon2020fedmix}. We also report the results of \emph{FedData}~\citep{zhao2018federated}  for reference but it is not considered as a baseline since raw data are exposed. All baselines are hyper-parameter optimized to report their best performances. The specific setting can be found in Appendix~\ref{Sec:Apen_C3}.

For all methods, we use the Adam optimizer for local training with an initial learning rate of  $10^{-3}$  and decay it by $2\%$  per communication round until  $10^{-5}$. The size of the local batch is $32$ and the number of local epochs is set to $5$. $10\%$ of clients are randomly sampled at each round. We run $100$ communications and report the best accuracy as the final result.   For all settings, we report the mean and standard deviation of the accuracy from five runs with different random seeds.  For \textit{FLea}, without particular mention, we use $\beta\sim  Beta(2,2)$ for Eq.~(\ref{mixup}), and $\lambda_1=1, \lambda_2=3$ for the loss in Eq.~(\ref{Eq:loss}), and extract the activations after several conventional layers of the model (as specified in Appendix~\ref{Sec:Apen_C2}). We only share features from $\alpha=10\%$ of the local data.

 \subsection{Performance comparison (RQ1)}\label{sec:results} 
The overall accuracy achieved by \textit{FLea} and baselines are compared in Table~\ref{table:result_all}. Across various model architectures (x$3$) and different levels of local data skewness (x$3$) and scarcity (x$2$), \textit{FLea} constantly outperforms the state-of-the-art baselines and significantly reduces the gap compared to \textit{FedData} which shares the raw data. In $13$ out of the studied $18$ scenarios, \textit{FLea} presents an improvement of over $5\%$, and among those, in $5$ cases, the improvement is more than $10\%$. These demonstrate the superiority and generality of \textit{FLea} in addressing the challenges caused by label skew and data scarcity.

\begin{figure}[t]
\vspace{-45pt}
  \begin{center}
 \includegraphics[width=0.46\textwidth]{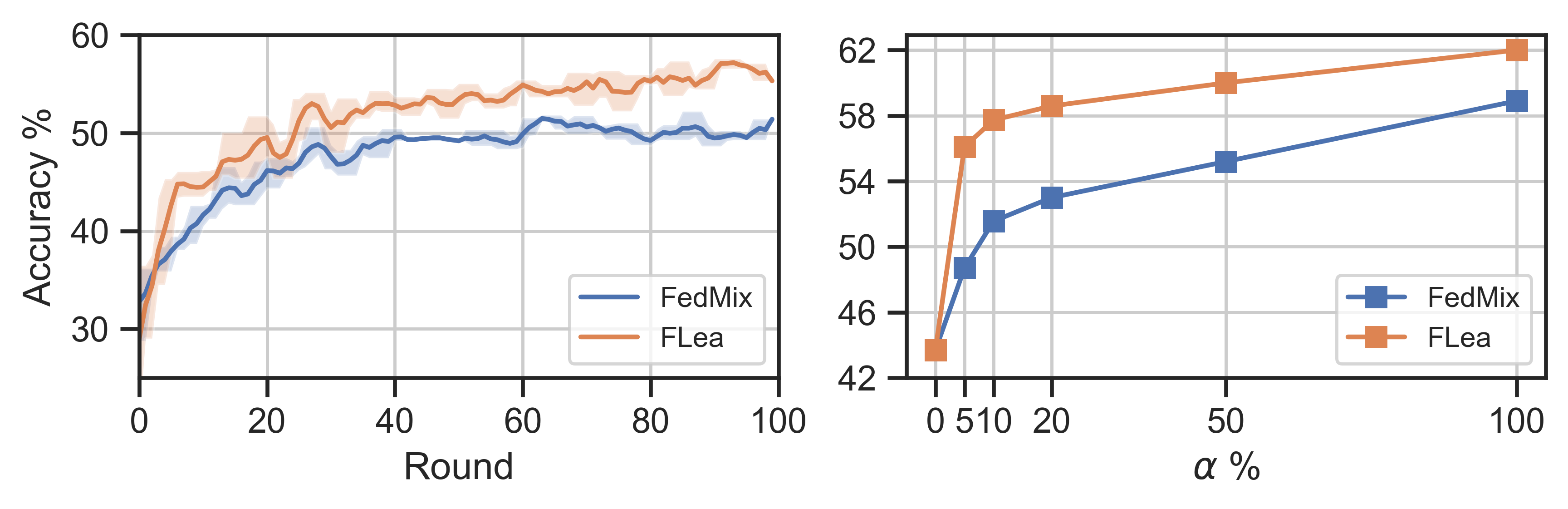}
  \end{center}
  \vspace{-15pt}
  \caption{Comparison between \textit{FedMix} and \textit{FLea} using UrbanSound$8$K with $Qan(3), \bar{|\mathcal{D}_k|}=100$.}
    \label{fig:round}
\vspace{-12pt}
\end{figure}

It can also be observed that the data-augmentation-based baselines, particularly \textit{FedMix}, outperform the loss-based baselines in most cases, which again supports \textit{FLea}'s use of
some global information to aid local training in the presence of data scarcity. In \textit{FedMix}, sample aggregations are shared to protect data privacy, yet this inevitably harms the utility of the shared augmentations for the training of classification models. On the contrary, \textit{FLea} shares the intermediate layer activations, which are obfuscated to protect privacy while retaining useful information for classification. This explains why \textit{FLea} can consistently outperform \textit{FedMix}. \textit{FLea} is also more efficient than \textit{FedMix}. As illustrated in Figure~\ref{fig:round}, \textit{FLea} learns faster than \textit{FedMix} after the first $5$ rounds (cf. the left figure) and requires a smaller proportion of augmentations to be shared to achieve the same accuracy as \textit{FedMix} (cf. the right figure).

\definecolor{mypink}{rgb}{.99,.91,.95}
\definecolor{mygrey}{rgb}{.9,.9,.9}
\begin{table*}[t]
\centering
    \caption{Overall accuracy comparison. Accuracy is reported as $mean\pm std$ across five runs.  The best performance under each setting is highlighted in red and the SOTA baseline (*excluding \textit{FedData}) is in grey. \textcolor{red}{$\uparrow$} indicates a relative improvement of our method compared to the SOTA over $5\%$ and \textcolor{red}{$\uparrow \uparrow$} indicates a relative improvement over $10\%$.}
    \vspace{-10pt}
    \resizebox{0.92\textwidth}{!}
    {%
    \begin{tabular}{@{}ll||ccc||ccc||ccc@{}}
    \toprule
      && \multicolumn{3}{c||}{\textbf{CIFAR10}} & \multicolumn{3}{c||}{\textbf{\revise{UrbanSound8K}}} & \multicolumn{3}{c}{\textbf{\revise{UCI-HAR}}}\\   
      & Accuracy $\%$& $Qua(3)$ & $Dir(0.5)$ &$Dir(0.1)$ & $Qua(3)$ & $Dir(0.5)$ &$Dir(0.1)$ &$Qua(2)$ & $Dir(0.3)$ &$Dir(0.1)$
      \\ \midrule
    \parbox[t]{5mm}{\multirow{11}{*}{\rotatebox[origin=c]{90}{$\bar{|\mathcal{D}_k|}=100$}}}&FedAvg      &$30.25{\pm 1.33}$&$32.58{\pm 1.09}$&$20.46{\pm 2.15}$ 
                &$43.69{\pm 0.56}$ &$46.77{\pm 0.87}$ &$34.59{\pm 2.64}$
                &$66.99{\pm 0.87}$ &$65.78{\pm 0.34}$ &$48.43{\pm0.70}$\\   
    &FedProx     &$31.92{\pm 1.45}$&$32.01{\pm 1.25}$&$20.86{\pm 1.97}$ 
                &$38.45{\pm 0.48}$&$39.58{\pm 1.02}$ &$34.81{\pm 0.46}$
                &$68.32 {\pm 0.50 }$ &$67.75{\pm 0.41}$ &$58.35{\pm0.52}$\\
    &FedDecorr   &$31.12{\pm 1.57}$&$33.57{\pm 1.22}$&$21.34{\pm 1.59}$ 
                &$45.01{\pm 0.57}$ &$46.77{\pm 0.65}$ &$35.87{\pm 1.03}$
                &$69.12 {\pm 0.63 }$ &$66.68{\pm 0.43}$ &$57.05{\pm0.38}$\\
    &FedLC       &$32.05{\pm 1.60}$&$30.17{\pm 1.18}$&$18.82{\pm 2.01}$ 
                &$50.98{\pm 0.49}$ &$50.11{\pm 0.83}$ &$37.05{\pm 0.87}$
                &\cellcolor{mygrey}$71.69 {\pm 0.52 }$ &\cellcolor{mygrey}$70.57{\pm 0.38}$ &$62.57{\pm0.42}$\\
    &FedNTD      &$39.98{\pm 0.97}$&$39.82{\pm 0.86}$&$26.78{\pm 2.34}$ 
                &$49.80{\pm 0.45}$ &$51.09{\pm 0.97}$ &$36.53{\pm 0.99}$
                &$68.33{\pm 0.72}$ &\cellcolor{mygrey}$70.32{\pm 0.49}$ &$60.13{\pm0.51}$\\
    \cmidrule(r){2-11}
    &FedBR   &$31.66{\pm 1.07}$& $33.08{\pm 1.12}$& $20.98{\pm 2.54}$ 
            &$44.05{\pm 0.63}$ &$47.58{\pm 0.90}$ &$36.15{\pm 1.17}$
            &$67.54{\pm 0.68}$ &$69.15{\pm 0.40}$ &$59.87{\pm0.46}$\\
    &CCVR    &$35.95{\pm 1.63}$& $35.02{\pm 1.43}$& $24.21{\pm 2.67}$ 
            &$47.12{\pm 0.72}$ &$49.26{\pm 0.92}$ &$39.62{\pm 1.20}$
            &$70.17{\pm 0.49}$ &$68.87{\pm 0.51}$ &$60.28{\pm0.36}$\\
    &FedGen  &$32.32{\pm 1.21}$& $34.27{\pm 1.56}$& $22.56{\pm 2.89}$ 
            &$45.20{\pm 0.89}$ &$48.33{\pm 1.12}$ &$38.27{\pm 1.44}$
            &$70.58{\pm 0.61}$ &$69.32{\pm 0.60}$ &$60.07{\pm0.63}$\\            
    &FedMix  &\cellcolor{mygrey}$44.04{\pm 1.53}$&\cellcolor{mygrey}$45.50{\pm 1.88}$&\cellcolor{mygrey}$38.13{\pm2.06}$  
            &\cellcolor{mygrey}$51.56{\pm 0.59}$&\cellcolor{mygrey}$54.18{\pm 0.62}$&\cellcolor{mygrey}$43.35{\pm 0.72}$ 
            &$68.59{\pm 0.54}$ &$69.34{\pm 0.49}$ &\cellcolor{mygrey}$65.63{\pm0.47}$\\
            &FedData* &$54.64{\pm 1.02}$&$56.47{\pm 1.22}$&$55.35{\pm 1.46}$ 
            &$62.83{\pm 1.25}$&$64.45{\pm 0.76}$ &$61.11{\pm 0.98}$
            &$78.13{\pm 0.46}$ &$78.24{\pm 0.51}$ &$75.93{\pm0.34}$\\ \cmidrule(r){2-11} 
    &\textbf{FLea} &\cellcolor{mypink}$47.03{\pm1.01}$\textcolor{red}{$\uparrow$}&\cellcolor{mypink}$48.86{\pm1.43}$\textcolor{red}{$\uparrow$}&\cellcolor{mypink}$44.40{\pm1.23}$\textcolor{red}{$\uparrow\uparrow$}
    &\cellcolor{mypink}$57.73 {\pm 0.51}$\textcolor{red}{$\uparrow\uparrow$}&\cellcolor{mypink}$59.22 {\pm 0.78}$\textcolor{red}{$\uparrow$} &\cellcolor{mypink}$45.94 {\pm 0.77}$\textcolor{red}{$\uparrow$}
    &\cellcolor{mypink}$75.17{\pm 0.42}$&\cellcolor{mypink}$73.02{\pm 0.49}$&\cellcolor{mypink}$71.68{\pm 0.51}$\textcolor{red}{$\uparrow$}
    \\\midrule \midrule
    \parbox[t]{5mm}{\multirow{11}{*}{\rotatebox[origin=c]{90}{$\bar{|\mathcal{D}_k|}=50$}}}
    &FedAvg      &$27.72{\pm1.26}$&$26.92{\pm 1.31}$&$21.88{\pm 1.87}$
    &$39.35{\pm 0.60}$ &$43.98{\pm 0.89}$ & $ 31.21{\pm 1.62}$
    &$65.77{\pm 0.42}$ &$67.10{\pm 0.40}$ &$46.95{\pm 0.62}$\\   
    &FedProx     &$ 22.88{\pm 2.54}$&$ 24.47{\pm 2.17}$&$ 21.01{\pm 2.46}$
    &$39.05{\pm 0.56}$ &$42.21{\pm 0.76}$ &$32.85{\pm1.22}$   
     &$69.18{\pm 0.41}$ &$68.28{\pm 0.45}$ &$59.97{\pm 0.46}$\\
   &FedDecorr   &$26.45{\pm1.58}$&$25.57{\pm 1.84}$&$ 22.03{\pm 1.98}$
    &$39.67{\pm 0.58}$ &$44.23{\pm 0.95}$ & $33.67{\pm 1.34}$
     &$65.77{\pm 0.39}$ &$68.57{\pm 0.51}$ &$55.54{\pm 0.49}$\\
    &FedLC       &$ 28.64{\pm 1.52}$&$ 26.36{\pm 1.47}$&$ 20.24{\pm1.68}$
    &$44.33{\pm 0.79}$&$45.15{\pm 0.80}$ &$39.87{\pm1.04}$
     &\cellcolor{mygrey}$70.63{\pm 0.49}$ &\cellcolor{mygrey}$71.34{\pm 0.45}$ &\cellcolor{mygrey}$63.67{\pm 0.52}$\\
    &FedNTD      &$32.92{\pm1.43}$&$34.64{\pm 1.52}$&$30.13{\pm1.67}$
    &$42.21{\pm 0.63}$ &$48.63{\pm 0.78}$ &$40.15{\pm1.22}$
     &$65.64{\pm 0.38}$ &$67.16{\pm 0.43}$ &$59.93{\pm 0.46}$\\
    \cmidrule(r){2-11}
    &FedBR  &$30.25{\pm1.45}$&$30.32{\pm1.32}$&$28.52{\pm 1.56}$
    &$41.15{\pm 0.70}$ &$44.37{\pm 0.82}$ &$34.89{\pm1.36}$
     &$66.98{\pm 0.43}$ &$68.23{\pm 0.49}$ &$57.25{\pm 0.52}$\\
    &CCVR    &$34.01{\pm1.89}$&$35.12{\pm1.34}$&$33.26{\pm1.56}$
    &$44.05{\pm 0.87}$ &$46.68{\pm 0.83}$ &$36.80{\pm1.37}$
     &$65.24{\pm 0.50}$ &$70.15{\pm 0.46}$ &$60.26{\pm 0.57}$\\
    &FedGen  &$33.12{\pm 1.61}$&$31.89{\pm 1.59}$&$29.90{\pm 1.76}$
    &$40.89{\pm 0.72}$ &$44.54{\pm 0.81}$ &$35.78{\pm1.40}$
     &$68.27{\pm 0.64}$ &$69.82{\pm 0.41}$ &$59.13{\pm 0.45}$\\
    &FedMix  &\cellcolor{mygrey}$38.14{\pm 1.12}$&\cellcolor{mygrey}$39.87{\pm1.55}$&\cellcolor{mygrey}$ 36.87{\pm1.38}$
    &\cellcolor{mygrey}$46.55{\pm 0.81}$&\cellcolor{mygrey}$50.00{\pm 0.92}$&\cellcolor{mygrey}$42.27{\pm 1.15}$  
     &$68.06{\pm 0.44}$ &$70.80{\pm 0.45}$ &$61.39{\pm 0.46}$\\
    &FedData* &$53.59{\pm 1.32}$&$53.02{\pm 1.18}$&$53.56{\pm1.64}$
    &$60.31{\pm 0.82}$&$60.48{\pm 0.91}$&$59.67{\pm 1.55}$
     &$76.42{\pm 0.38}$ &$76.45{\pm 0.47}$ &$75.46{\pm 0.47}$\\\cmidrule(r){2-11}
    &\textbf{FLea} &\cellcolor{mypink}$41.98{\pm1.26}$\textcolor{red}{$\uparrow\uparrow$}&\cellcolor{mypink}$42.01{\pm1.13}$\textcolor{red}{$\uparrow$}&\cellcolor{mypink}$37.69{\pm 1.65}$
    &\cellcolor{mypink}$54.35{\pm 0.80}$\textcolor{red}{$\uparrow\uparrow$}&\cellcolor{mypink}$55.68{\pm 0.87}$\textcolor{red}{$\uparrow\uparrow$}&\cellcolor{mypink}$45.05{\pm 1.32}$\textcolor{red}{$\uparrow$}
    &\cellcolor{mypink}$74.25{\pm 0.44}$\textcolor{red}{$\uparrow$}&\cellcolor{mypink}$73.98{\pm 0.46}$&\cellcolor{mypink}$ {66.57\pm0.45}$\\
\bottomrule
    \end{tabular}}
\label{table:result_all}
\vspace{-10pt}
\end{table*}

\subsection{Effects of key design choices (RQ2)}
\subsubsection{The role of augmentation.} 
\begin{figure}[t]
  \centering
  \vspace{-1pt}
  \includegraphics[width=0.38\textwidth]{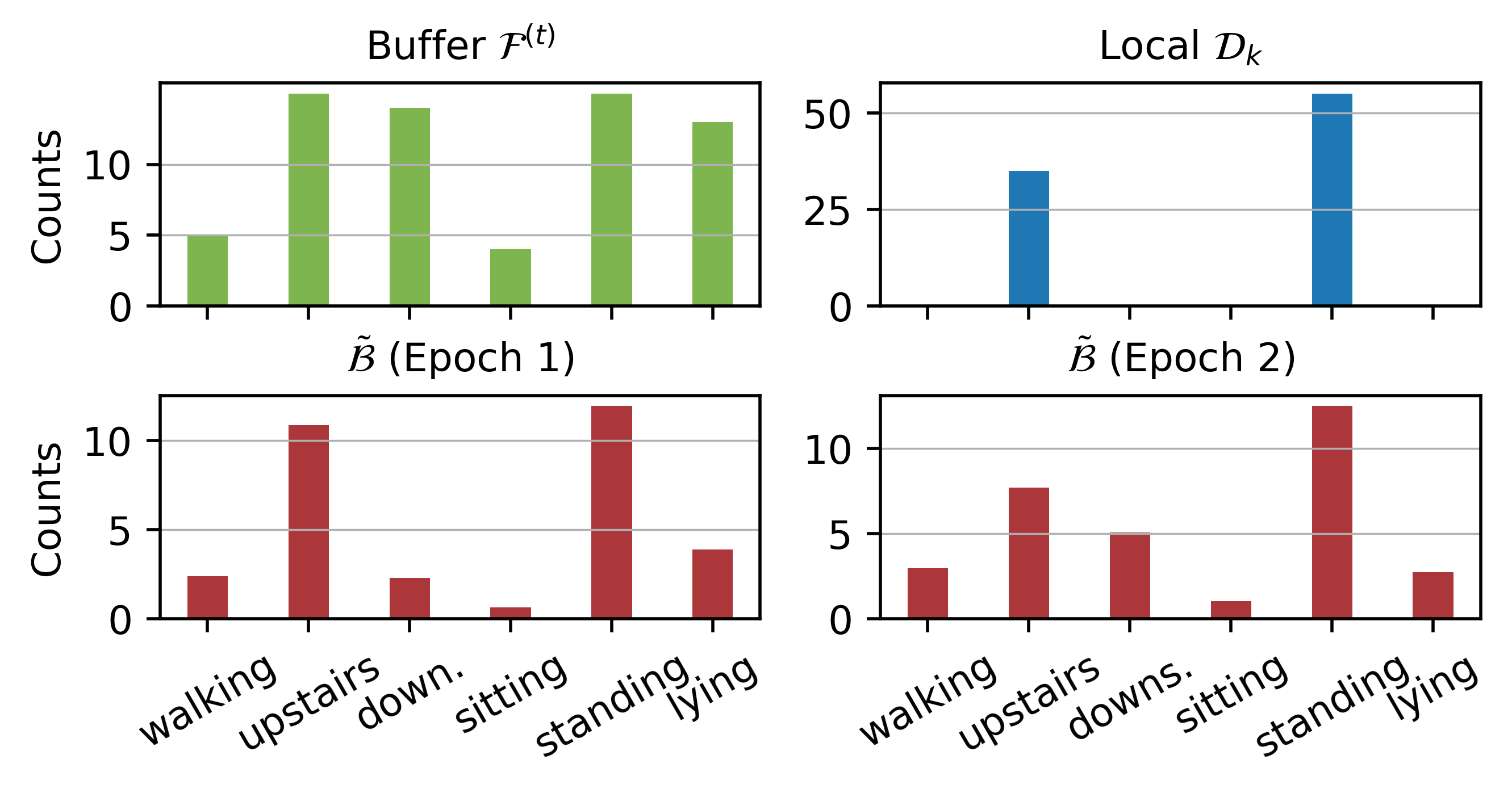}
  \vspace{-10pt}
  \caption{Class distributions for the global feature buffer $\mathcal{F}^{(t)}$, the local data $\mathcal{D}_k$, and augmented batch $\mathcal{\tilde{B}}$ in different epochs. The example is from UCI-HAR with $Qan(2), \bar{|\mathcal{D}_k|}=100$.}
  \label{fig:HAR_1}
\end{figure}

The feature buffer is a key component in \textit{FLea}, designed to compensate for the missing classes in the local dataset. Figure~\ref{fig:HAR_1} displays an example of the class distribution for the global feature buffer $\mathcal{F}^{(t)}$ and the local data $\mathcal{D}_k$ at round $t$. $\mathcal{F}^{(t)}$ contains features from all six classes, while $\mathcal{D}_k$ only has data from two classes.
As introduced in Sec.\ref{sec:aug}, a batch of features is sampled from $\mathcal{F}^{(t)}$ and $\mathcal{D}_k$, respectively. These features are then mixed up to create the augmented input $\mathcal{\tilde{B}}$, as shown in the red bar plots in Figure~\ref{fig:HAR_1}: they also contain all six classes. This explains how the feature buffer and mix-up augmentation help address the local label skew.

\begin{figure}[t]
  \centering
  \vspace{-12pt}
  \includegraphics[width=0.4\textwidth]{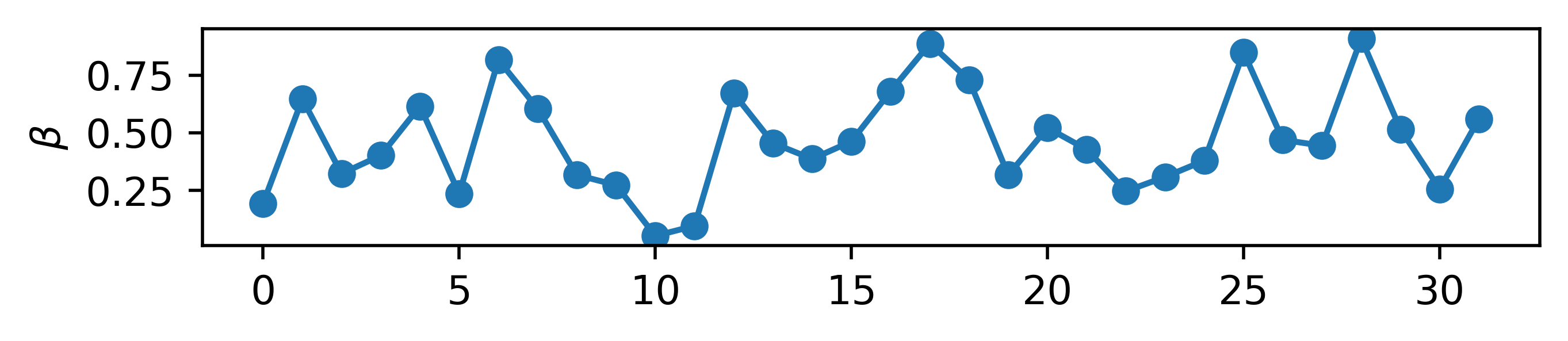}
  \vspace{-10pt}
  \caption{Value of $\beta$ for one batch (batch size $=32$).}
  \label{fig:HAR_2}
    \vspace{-15pt}
\end{figure}

Since we dynamically sample the weight $\beta$ for the mix-up augmentation, we can generate infinite combinations of global features and local features. An example of $\beta$ for one batch is given in Figure~\ref{fig:HAR_2}, and the corresponding class distribution for the augmented batch in different epochs are shown in Figure~\ref{fig:HAR_1}. It is worth noting that the two red bar plots in Figure~\ref{fig:HAR_1} are different, although they are generated using the same global and local features. Such diversity in the training data can significantly enhance the generalization of the local model, ultimately alleviating local overfitting as introduced in Sec.~\ref{Sec:DB} and improving the global model.

\subsubsection{Impact of hyper-parameters.} We now analyze how the hyper-parameters including the weight $\beta$ in Eq.~(\ref{mixup}) and weights $\lambda$ in the loss Eq.~(\ref{Eq:loss}) affect the performance of \textit{FLea}. We use UrbanSound$8$K under the setting of $Qua(3)$ and $\bar{|\mathcal{D}_k|}=100$ as an example for a detailed analysis.
We first look at the impact of $\beta$. In Figure~\ref{fig:parmater_beta}, we visualize the density function for  $\beta\sim Beta(a,a)$ in the left, and the model performance with varying hyper-parameter $a$ controlling the shape of the distribution in the right. From the results, we can observe that training with augmentation regardless of the viable of $\beta$ outperforms training without mix-up (\textit{FedAvg}), while a choice of $a$ ranging from $1.5$ to $5$ yields the best performance. Thus we set $a=2$ in our experiments.
 
Regarding the weights in the loss function, we first set $\lambda_2=0$ (without obfuscating the features) and search the value for $\lambda_1$. As shown in Figure~\ref{fig:hyper_loss} (the left one), we found that  $\lambda_1>1$ can improve the performance compared to that without the distilling loss ($\lambda_1=0$), but if the weight is too large ($\lambda_1>4$) it harms the performance. The pattern is similar with other $\lambda_2$, and thus we informally use $\lambda_1=1$ for all experiments. With $\lambda_1=1$, we further study how $\lambda_2$ impacts the trade-off between privacy preservation (reflected by the reduced correlation) and the feature utility (reflected by the model accuracy), as shown in  Figure~\ref{fig:hyper_loss} (the right one). Enlarging $\lambda_2$ can significantly enhance privacy protection (referring to the increasing $1-\bar{c}$) but decreases the final performance. We finally use $\lambda_2=3$ when the $\bar{c}$ reduces to about $0.72$ while maintaining a strong accuracy of about $57\%$. We also suggest future applications using $2\sim 6$ for the trade-off.

\begin{figure*}[t]
 \vspace{-10pt}
  \centering
   \subfigure[Feature exposure. ]{
  \includegraphics[width=0.175 \textwidth]{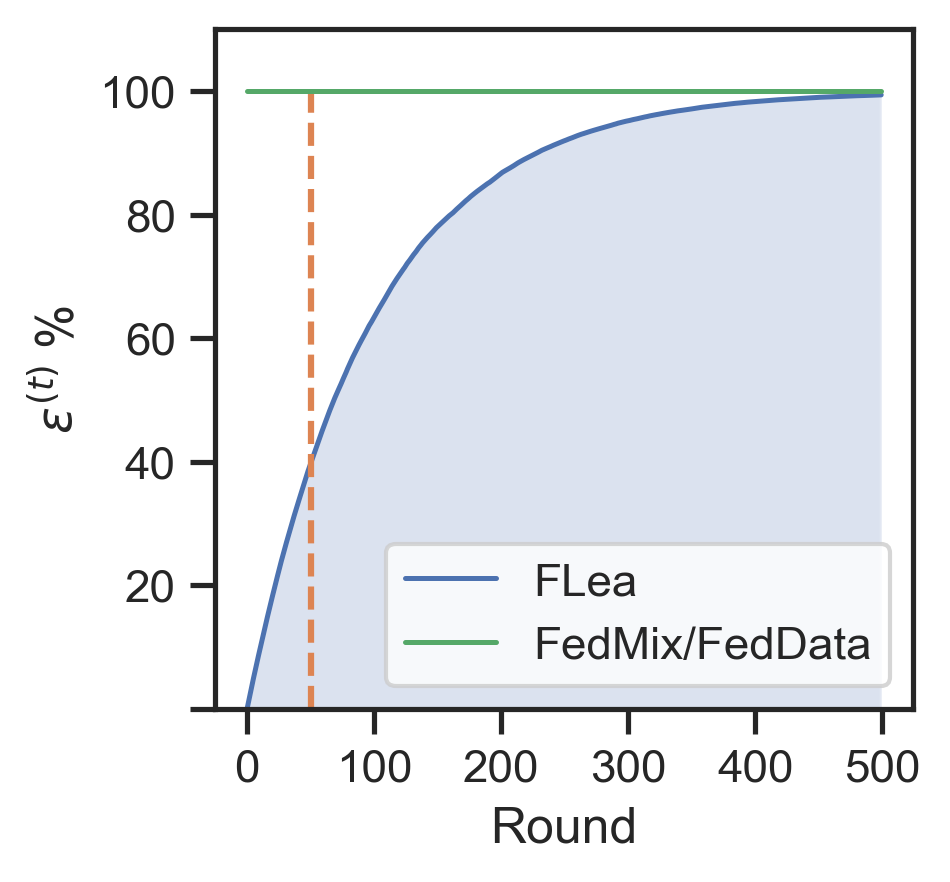}\label{fig:fea_e}}
   \subfigure[Correlation. ]{
  \includegraphics[width=0.17 \textwidth]{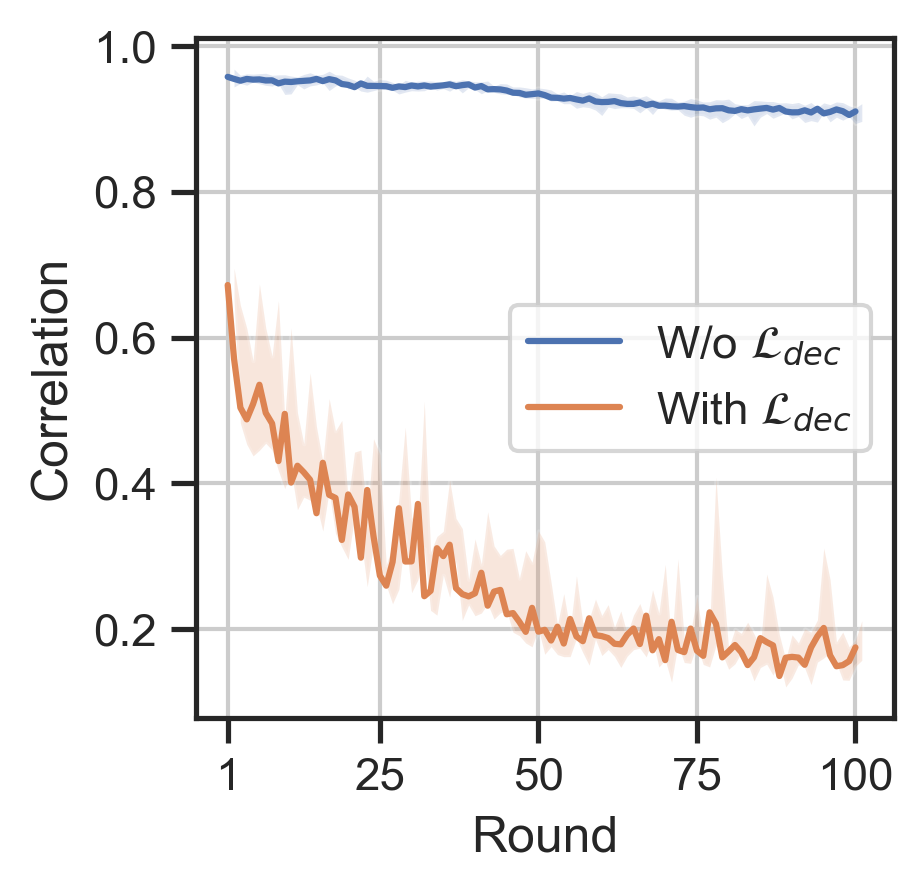}\label{fig:corr}}
    \subfigure[Reconstruction training error. ]{
  \includegraphics[width=0.29 \textwidth]{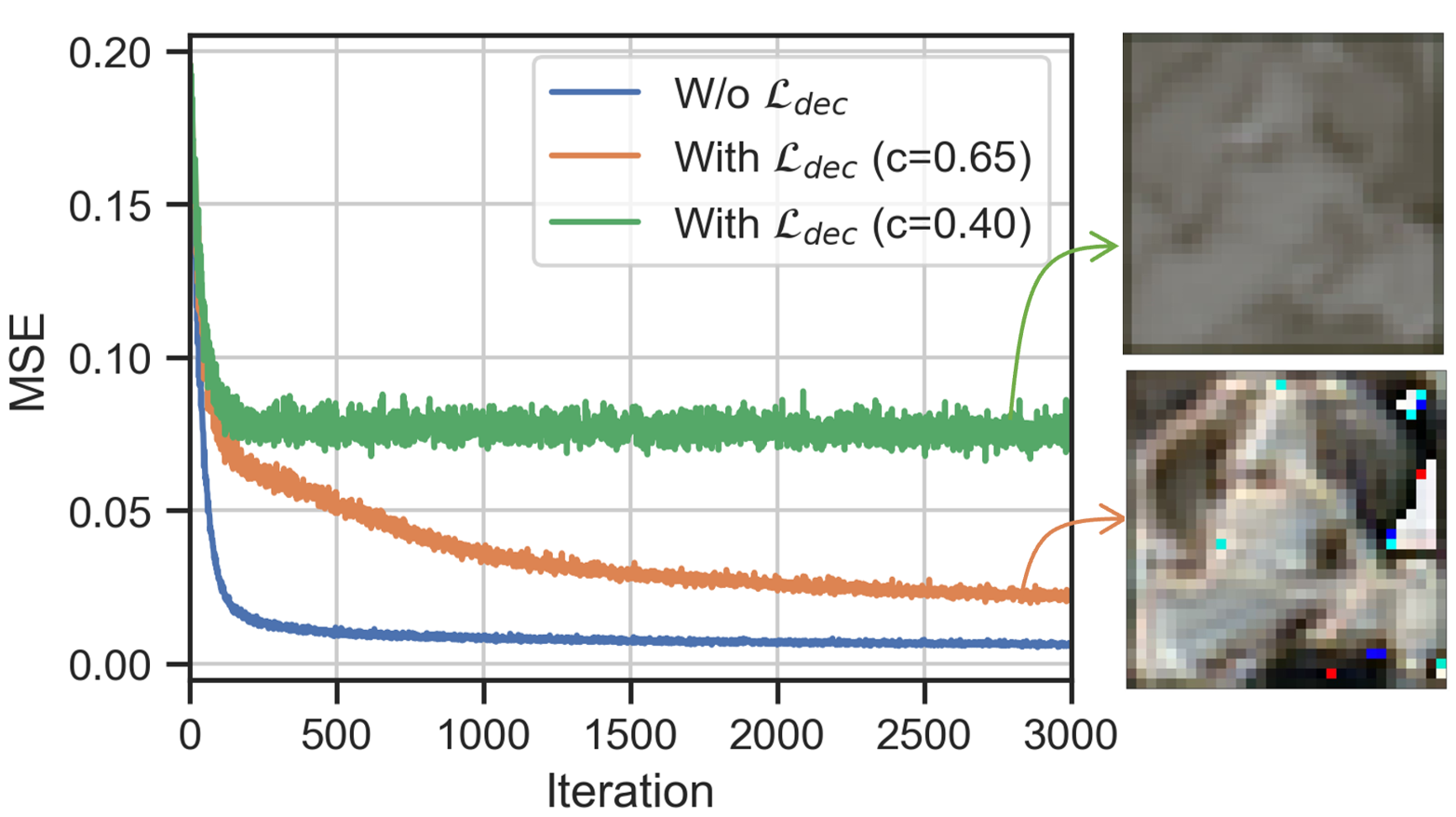}\label{fig:recons}}
      \subfigure[Context detection accuracy.]{
  \includegraphics[width=0.29 \textwidth]{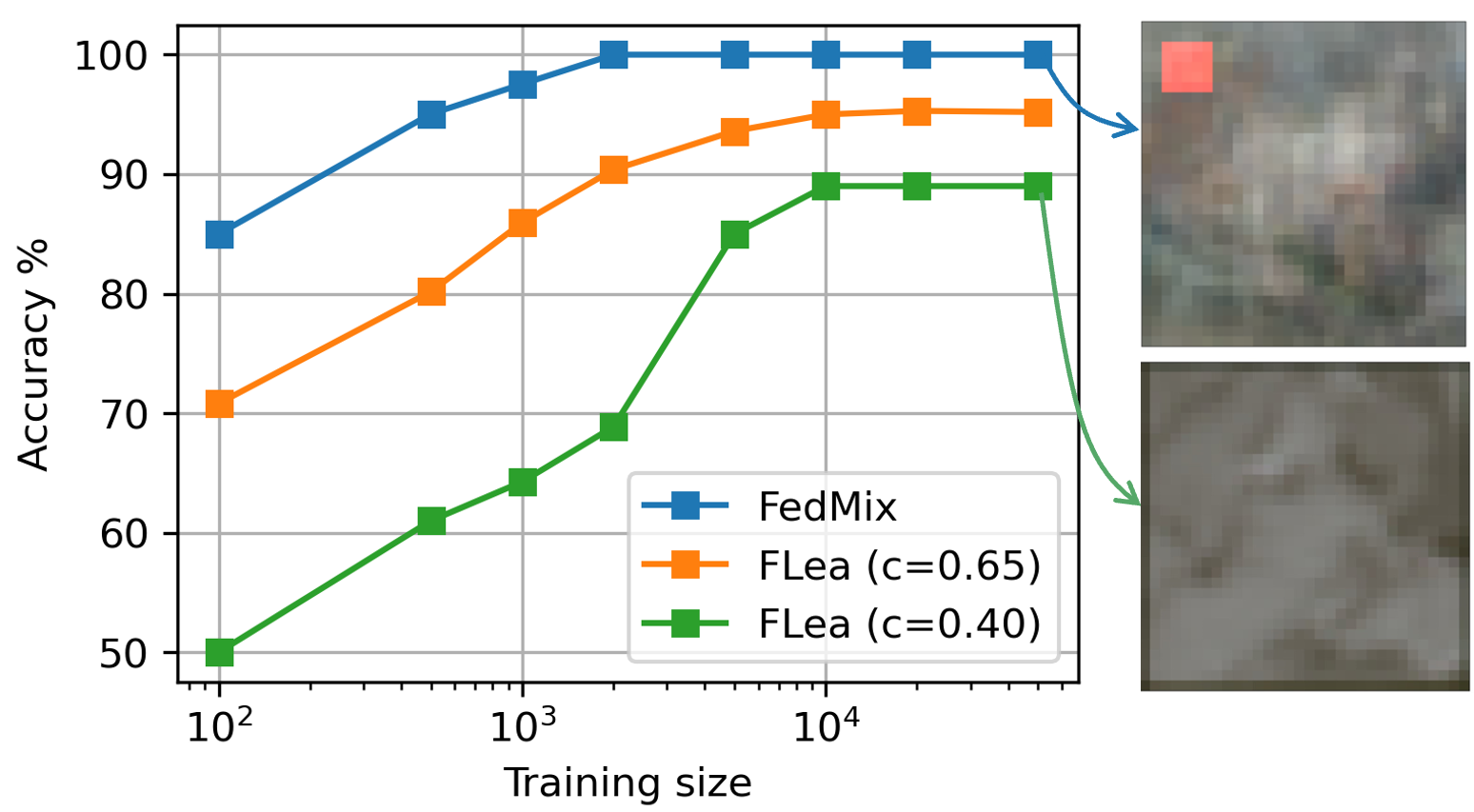}\label{fig:recall}}
  \vspace{-10pt}
  \caption{The effectiveness of privacy protection. $c$ is short for the expected correlation in (b). We show the reconstruction and context detection performance for $c=0.65$ (the $1^{st}$ round) and $c=0.40$ (the $10^{th}$ round).} \label{fig:mm}
  \vspace{-10pt}
\end{figure*}

\begin{figure}[t]
  \centering
  \includegraphics[width=0.4\textwidth]{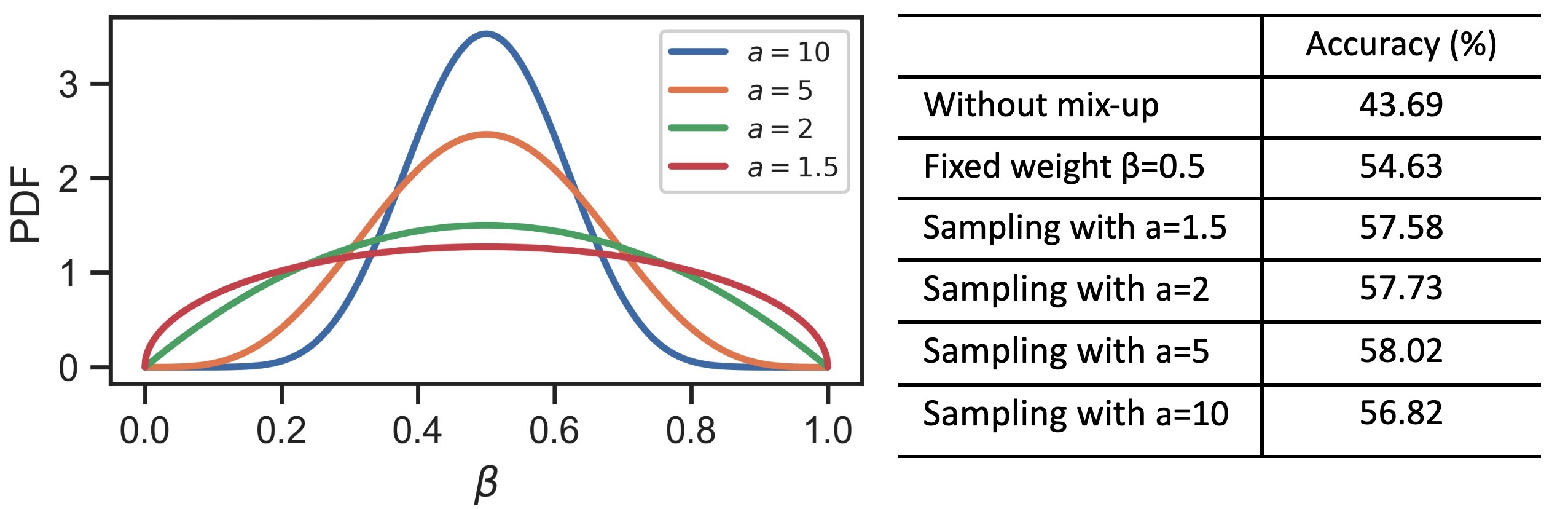}
  \vspace{-10pt}
  \caption{Beta distribution $\beta\sim Beta(a,a)$ and the model performance with varying $a$.}
  \label{fig:parmater_beta}
\end{figure}

  \vspace{-2pt} 
\subsection{Privacy analysis (RQ3)}
Now, we show \textit{FLea} can protect the privacy of the shared features from the following aspects: \textit{i)} reducing exposure by only sharing features with a fraction of clients, \textit{ii)} preventing data reconstruction, and \textit{iii)} impeding context information identification as introduced in Sec.~\ref{Sec:2.4}. We use CIFAR10 ($Qua(3), |\bar{\mathcal{D}}_k=100|$) as the example to demonstrate that \textit{FLea} is more privacy-preserving than its counterpart \textit{FedMix} and \textit{FedData}.


\noindent\textbf{Reducing feature exposure.}
As described in Sec.~\ref{Sec:3.4}, in \textit{FLea}, features from only a small fraction of local data are shared among the selected $10\%$ of the clients each round (clients in round $t-1$ share with clients in $t$).
To quantify the feature exposure, we define a feature exchange matrix $\xi \in \mathbb{R}^{|K| \times |K|}$ ($|K|$ is the number of total clients). 
$\xi_{i,j}^{(t)} = 1$ denotes client $i$ and $j$ have exchanged features for at least once until (including) $t$-th round, otherwise $\xi_{i,j}^{(t)} = 0$. The feature exposure is measured by $\epsilon^{(t)} = \sum_{i,j} \xi_{i,j}^{(t)}/|K|^2$ ( $0\leq\epsilon^{(t)}\leq1$), and a smaller $\epsilon^{(t)}$ is better.  As \textit{FedData} and \textit{FedMix} gather data or data averages and broadcast them to all clients before local model training, $\epsilon^{(t)}=100\%$ consistently. We illustrate $\epsilon^{(t)}$ for \textit{FLea} in Figure~\ref{fig:fea_e}: the exposure of \textit{FLea} grows slowly. In our experiments, the model converges within $50$ rounds (c.f. the learning curve in Figure~\ref{fig:round}), by when $\epsilon^{(t)} \leq40\%$.  Therefore, feature exposure is reduced by \textit{FLea}. 

Moreover, feature exposure is not equivalent to privacy leakage. Based on the reduction of distance correlation between raw data and learned activations during training (as shown in Figure~\ref{fig:corr}), \textit{FLea} is resilient to data reconstruction and context identification attacks. We construct test beds for a quantitative evaluation (detailed setup can be found in Appendix~\ref{sec:attacks}) and report the results below.

\begin{figure}[t]
  \centering
  \vspace{-12pt}
  \includegraphics[width=0.38\textwidth]{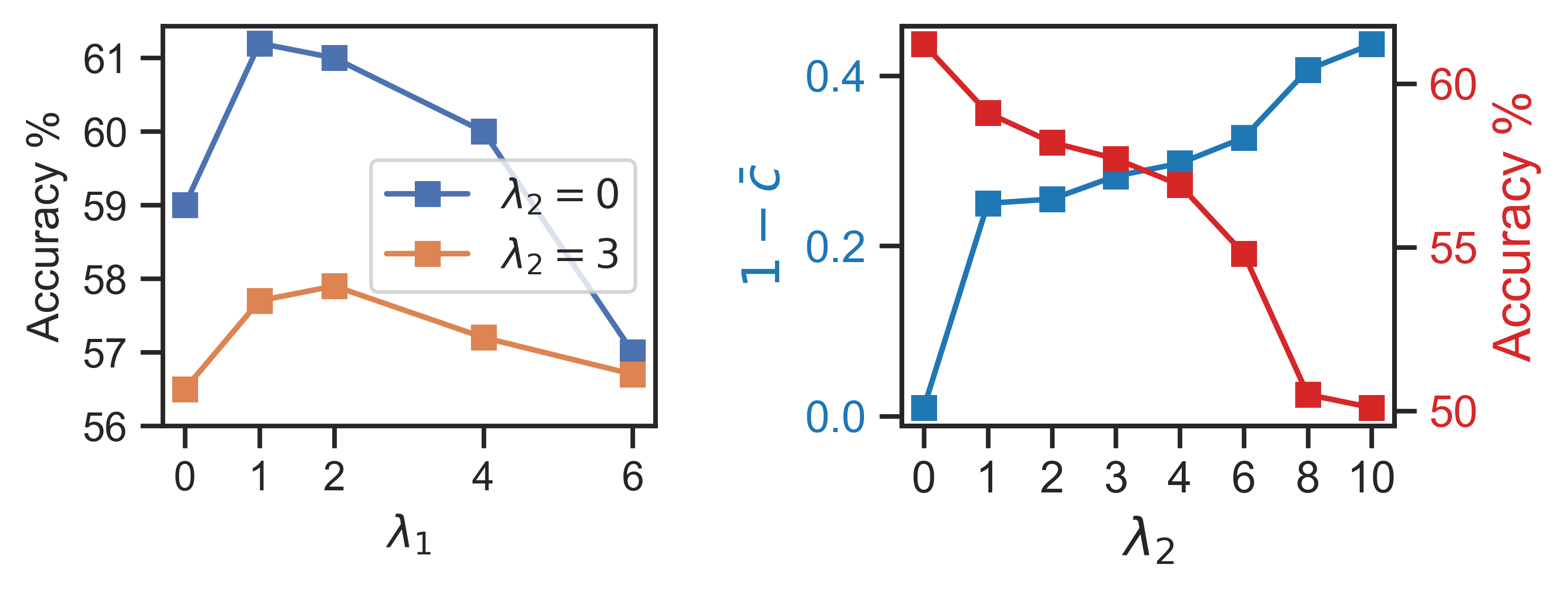}
  \vspace{-10pt}
  \caption{Impact of $\lambda_1$ and $\lambda_2$. $\bar{c}$ denotes the expectation of the distance correlation between activations and source data.}
  \label{fig:hyper_loss}
  \vspace{-12pt}
\end{figure} 

\vspace{1pt}
\noindent\textbf{Preventing data reconstruction}. We built an attacker model by using the activations from the FL model with and without reducing the distance correlation, respectively~\cite{vepakomma2020nopeek}. The reconstruction error (MSE) for training the attacker is presented in Figure~\ref{fig:recons}. It can be observed with $\mathcal{L}_{dec}$, MSE can never be reduced to the value of $0.01$ achieved with normal training without $\mathcal{L}_{dec}$, which suggests the data cannot be accurately reconstructed. To further illustrate this, let's look at an example: Assume one client is selected to share the activation of an image, e.g., the dog image in Figure~\ref{fig:datashared}(a), in a certain communication round (when $c=0.4$),  the attacker tries to reconstruct the image from the shared activation. Our experiment shows the recovered image ends up with Figure~\ref{fig:datashared}(c*) (also shown in Figure~\ref{fig:recons}). The original attribute, i.e., the color distribution cannot be recovered and thus privacy is preserved. 

\vspace{1pt}
\noindent\textbf{Impeding context identification}. 
Our baseline \textit{FedMix} also demonstrates to be resilient to data reconstruction attacks because raw data are aggregated before sharing~\cite{yoon2020fedmix}. However, \textit{FedMix} can release context information, while \textit{FLea} can increase the difficulty of identifying the context. To quantify this, we constructed a context identification attacker, assuming that the context targeted for attack is represented by a colored square in the image (simulating a malfunctioning camera, see Figure~\ref{fig:datashared}(a*))). We employed a binary classifier to predict the presence of the context. Half of the CIFAR10 dataset was augmented with the marker for training and testing the attacker. The identification accuracy achieved with varying amounts of training samples is summarized in Figure~\ref{fig:recall}. The results indicate that \textit{FLea} requires significantly more training data (hundreds of times) than \textit{FedMix} to achieve comparable identification accuracy. For instance, to attain a $90\%$ accuracy rate, \textit{FedMix} needs approximately $300$ training samples, whereas \textit{FLea} ($c=0.4$) demands $10,000$ samples. It's important to note that real-world scenarios pose greater challenges, as the context pattern may not be explicit and the attacker might not have access to extensive training data, resulting in reduced attacking accuracy. In such scenarios, \textit{FLea} can effectively safeguard context privacy.

\vspace{-1pt}
\section{Conclusions}

We proposed \emph{FLea}, a novel approach to tackle scarce and label-skewed data in FL. Feature augmentation is employed to mitigate over-fitting and local drift simultaneously. Extensive experiments demonstrate that \emph{FLea} remarkably outperforms the state-of-the-art baselines while mitigating the privacy risk associated with feature sharing. 
In practice, \textit{FLea} can introduce some additional overheads, such as increased communication and storage requirements, due to feature sharing. We leave the task of improving efficiency for future work. To enhance privacy, \textit{FLea} can be combined with other methods like differential privacy~\cite{abadi2016deep} and homomorphic encryption~\cite{fang2021privacy}.
For applications such as healthcare, protecting the label distribution may be necessary, which we did not address in this paper. We anticipate that our work will inspire further investigations into comprehensive studies on feature sharing in low data regimes.

\vspace{-1pt}
\section*{Acknowledgment}

This work was supported by  European Research Council Project 833296 (EAR) and 805194 (REDIAL).
We also thank Prof. Nicholas Lane and Lorenzo Sani for their insightful discussions.

\bibliographystyle{ACM-Reference-Format}
\bibliography{sample}

\newpage 
\appendix
\section*{Appendix}

\section{Experimental setup for Sec.~2.3}\label{Sec:Apen_A}
\textbf{Federated Learning Setup.} In Sec.~\ref{Sec:scarcity}, we employ CIFAR10 for an empirical comparison. We conducted three groups of experiments to simulate different levels of data scarcity, as introduced below.
\begin{itemize}
    \item \textbf{(a)} The training set of CIFAR10 is uniformly distributed to $10$ clients, resulting in each local dataset having a size of $5000$ ($|\mathcal{D}_k|=5000$) and $10$ classes (IID). More specifically, each local dataset has $500$ samples per class.
    \item \textbf{(b)} CIFAR10 is uniformly distributed to $100$ clients. Thus each local dataset has a size of $500$ ($|\mathcal{D}_k|=500$), $10$ classes (IID) with each class containing $50$ samples. 
    \item \textbf{(c)} CIFAR10 is uniformly distributed to $500$ clients. Each local dataset have a size of $100$ ($|\mathcal{D}_k|=100$), $10$ classes (IID) with each class containing $10$ samples. 
\end{itemize}

For classification, we employ MobileNet\_V2, which has $18$ blocks consisting of multiple convolutional and pooling layers~\citep{sandler2018mobilenetv2}. We use the Adam optimizer for local training with an initial learning rate of $10^{-3}$ and decay it by $2\%$ per communication round until $10^{-5}$. For (a), all clients will participate in the training in each round, while for the other groups, we will randomly select $10\%$ of the clients for each round.
The size of the local batch is $64$, and we run $10$ local epochs for each group. We run 100 communication rounds for all groups to ensure global convergence.

\textbf{Experimental setup for Figure 1:} To compare the performance of existing methods with , we use CIFAR$10$ dataset
and report the classification accuracy of the global model based on the global testing set. We compare \textit{FedAvg} with loss-based methods such as \textit{FedDecorr} and \textit{FedNTD}, as well as data augmentation-based methods like \textit{FedMix} and \textit{FedData}. They are the most representative methods in each category. 
\textit{FedMix} is implemented by averaging every $10$ samples and sharing the result globally. The shared averaged data is then combined with local data according to a Beta distribution (with the $a=2$) for local training.
In the case of \textit{FedData}, we collect $10\%$ of the data (randomly chosen) from each client and share it globally, in the first communication round.
To simulate varying scarcity levels, we split the CIFAR$10$ training set (comprising $50,000$ samples in total) into $5000$, $500$, and $100$ training samples on average per client, which ends up with $10$, $100$ and $500$ clients finally.
Other settings are the same with the main experiments as introduced in Sec.~\ref{Sec: setup}.

\textbf{Experimental setup for Figure 2:} \revise{DB score}~\citep{davies1979cluster} is defined as the average similarity measuring each cluster with its most similar cluster, where similarity is the ratio of within-cluster distances to between-cluster distances. Thus, clusters which are farther apart and less dispersed will result in a better score. The minimum score is zero, with lower values indicating better clustering. To calculate the score for features, we use the ground-true class labels as cluster labels, and use Euclidean distance between features to measure the similarity. 

For a fair comparison, the local training for all clients starts from a same global status with an accuracy of $40\%$. The features of the testing set from the initial global model present a DB of $4.8$. We run one communication round and report the performance for the global model. In this round, for $|\mathcal{D}_k|=5000$ we aggregate $10$ clients while for $|\mathcal{D}_k|=100$ we aggregate $50$ clients, so that the total samples used for model training are kept unchanged.  For $|\mathcal{D}_k|=100+1000$ group, we additionally give the selected $50$ clients $1000$ samples (gathered in the first round) to aid local training. 
In Figure~2, for local models, we report the averaged DB across clients.

\section{Notations}\label{Sec:Apen_B}

\textbf{Beta Distribution.} The probability density function (PDF) of the Beta distribution is given by,
\begin{equation}
    p(\beta; a,b) = \frac{\beta^{a-1} (1-\beta)^{(b-1)} }{N},
\end{equation}
where $N$ is the normalizing factor and $\beta\in[0,1]$. In our study, we use a symmetrical distribution so that we choose $a=b$ and herein,   $p(\beta)= \frac{1}{N} (\beta (1-\beta)) ^{a-1}$.

\textbf{De-correlation Loss.} We employ the following formulation to quantify the correlation between the activation and the source data.
\begin{equation} 
\mathcal{L}_{dec}(\mathcal{B}) =  \frac{\nu^2(x,f)}{\sqrt{\nu^2(x,x)\nu^2(f,f)}},
\end{equation}  
where $\nu^2(,)$ denotes the squared distance. Specifically, $\nu^2(x,f) = \frac{1}{|\mathcal{B}|^2}\sum_{i,k}^{|\mathcal{B}|} \hat{E_x}[i,k]\hat{E_f}[i,k]$, and $\nu^2(x,x) = \frac{1}{|\mathcal{B}|^2}\sum_{i,k}^{|\mathcal{B}|} \hat{E_x}[i,k]\hat{E_x}[i,k]$.  $E_x$ is the Euclidean distance matrix for $x\in \mathcal{B}$, i.e., $E_x[i,k]=||x_i - x_k||^2$), and similarly $E_f[i,k]=||f_i - f_k||^2$. They are then double-centered to $\hat{E_X}$ and $\hat{E_F}$. To do this, we leverage the centering matrix of size $n=|\mathcal{B}|$ is defined as the $n$-by-$n$ matrix:
\begin{equation} 
C = I - \frac{1}{n}J,
\end{equation}
$I$ is the identity matrix of size $n$ and $J$  is an $n$-by-$n$ matrix of all $1$'s. Given this, $\hat{E_x}$ can be derived by,
\begin{equation}\label{eq:double_x} 
\hat{E}_x= C \times E_x \times C.
\end{equation}
$\hat{E_f}$ can be derived by,
\begin{equation}\label{eq:double_f} 
\hat{E}_f= C \times E_f \times C.
\end{equation}
Such double centralization can remove the effects of row and column means in the Euclidean distance matrices, making the distance more amenable to analyzing the correlation between $x$ and $f$.

\section{Details of Experiments}\label{Sec:Apen_C}

 \begin{table}[t]
    \caption{Statistics of the used datasets.}
    \centering
    \resizebox{0.48\textwidth}{!}
    {
    \begin{tabular}{l||ccccc }
    \toprule
         Data    &Size                  & $\#$Class &$\#$Training &$\#$Testing  &Model\\ \midrule
         CIFAR10 & $32\times 32\times 3$ &$10$       &50,000       &10,000       &MobiNet$\_$V2 \\
         UrbanSound$8$K&$2\times 64\times 344$ &$10$&6,986 &1,732 &AudioNet \\
         UCI-HAR &$128\times 3$&6&7,352&2,947&HARNet \\
    \bottomrule
    \end{tabular}
    }
    \label{tab:data} 
\end{table}

\subsection{Data Distribution}\label{Sec:Apen_C1}

An overview of the data we used is presented in Table~\ref{tab:data}. More details are given below.

\begin{figure*}[h]
  \centering
  \subfigure[Cifar$10$, $|K=1000|$, Qua($3$).]{
  \includegraphics[width=0.8\textwidth]{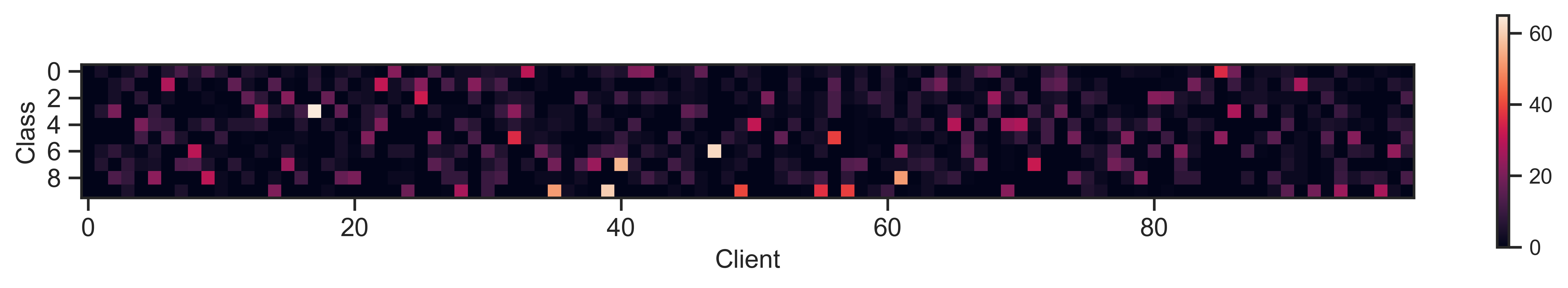}\label{data:cifar10}}

    \subfigure[UrbanSound$8$K, $|K=140|$, Dir($0.5$).]{
  \includegraphics[width=0.9\textwidth]{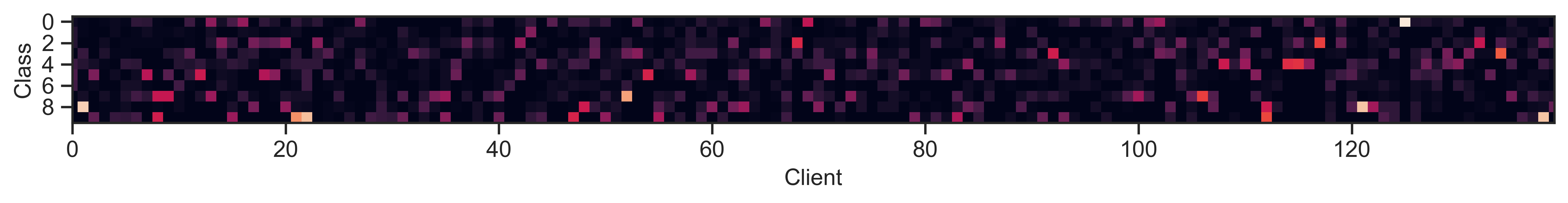}\label{data:audio}}
   
   \subfigure[UCI-HAR, $|K=150|$, Dir($0.1$).]{
  \includegraphics[width=0.99\textwidth]{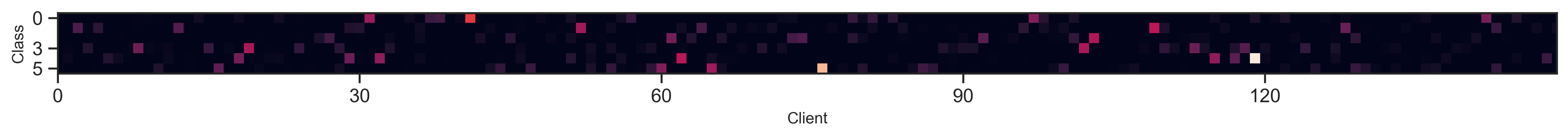}\label{data:audio}}
  \caption{Training data split for $\bar{|\mathcal{D}_k|}=50$. The local data residing on each client is scarce and label-skewed. }
  \label{fig:data}
\end{figure*}

\textbf{Image data:} We test our algorithm on CIFAR$10$~\citep{cifar10}. We distribute CIFAR$10$ training images (containing $50,000$ samples for $10$ classes) to $|K|=500$ and $|K|=1000$ clients and use the global CIFAR$10$ test set (containing $1,000$ samples per class) to report the accuracy of the global model. We show the data splits for the first $100$ clients when $|K|=5000$ in Figure~\ref{data:cifar10}.

\textbf{Audio data}: We also test \textit{FLea} using UrbanSound$8$K dataset~\cite{salamon2014dataset}. This dataset contains $8,732$ labeled sound excerpts ($\leq4s$) of urban sounds from 10 classes: air conditioner, car horn, children playing, dog bark, drilling, engine idling, gunshot, jackhammer, siren, and street music.  For experiments, we randomly hold out $20\%$ (about $1700$ samples) for testing and distribute the rest (about $7000$ samples) to $K$ clients for training.  We split the data into $|K|=70$ and $|K|=140$ folds, leading to an average local data size of the order of $100$ and $50$, respectively. The distribution for $|K=140|$, Dir($0.5$) is presented in Figure~\ref{data:audio}.

\textbf{Sensory data:} Sensory data is another modality we experiment with for its popularity, which can collected by wearable and mobile devices. UCI-HAR~\cite{misc_human_activity_recognition_using_smartphones_240} is a commonly used human activity recognition benchmark. It was collected by a waist-mounted smartphone with an embedded accelerometer. Six activities including walking, walking upstairs, walking downstairs, sitting, standing, and lying were recorded. We employ \textit{HARNet} which comprises 4 convolutional layers to recognize those activitie~\cite{teng2020layer}.  We split the data into training and testing sets, then distributed the training set to $75$ and $150$ clients to simulate different levels of scarcity, respectively. We show the data splits for $150$ clients when $\bar{|\mathcal{D}_k|}=50$ in Figure~\ref{data:cifar10}. 

\begin{figure}[h]
  \centering
  \subfigure[With label skew.]{
  \includegraphics[width=0.23\textwidth]{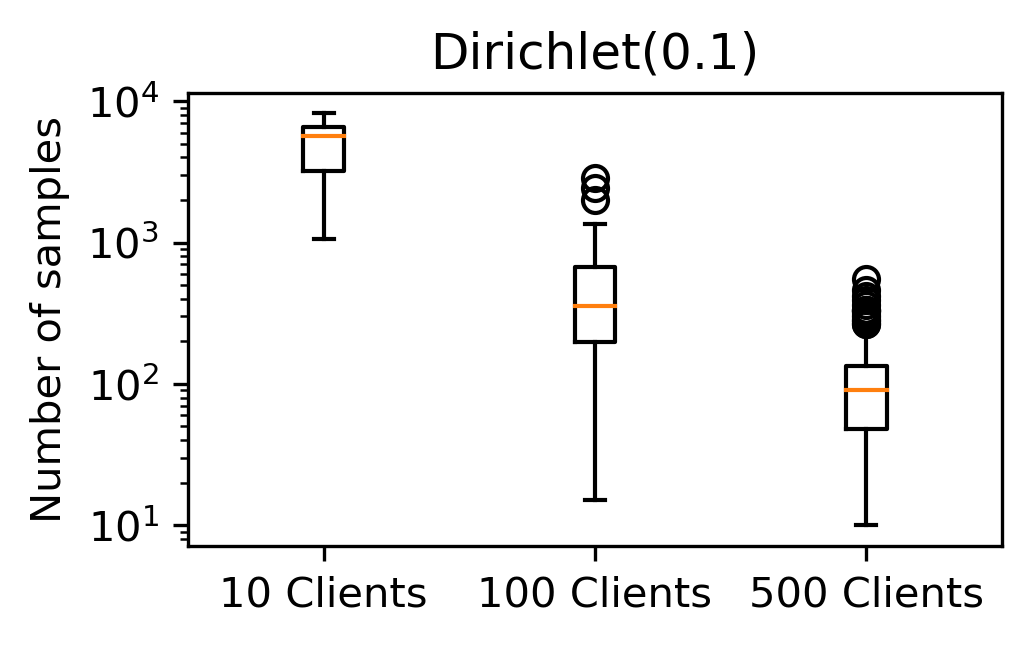}}
  \subfigure[Without label skew.]{
  \includegraphics[width=0.23\textwidth]{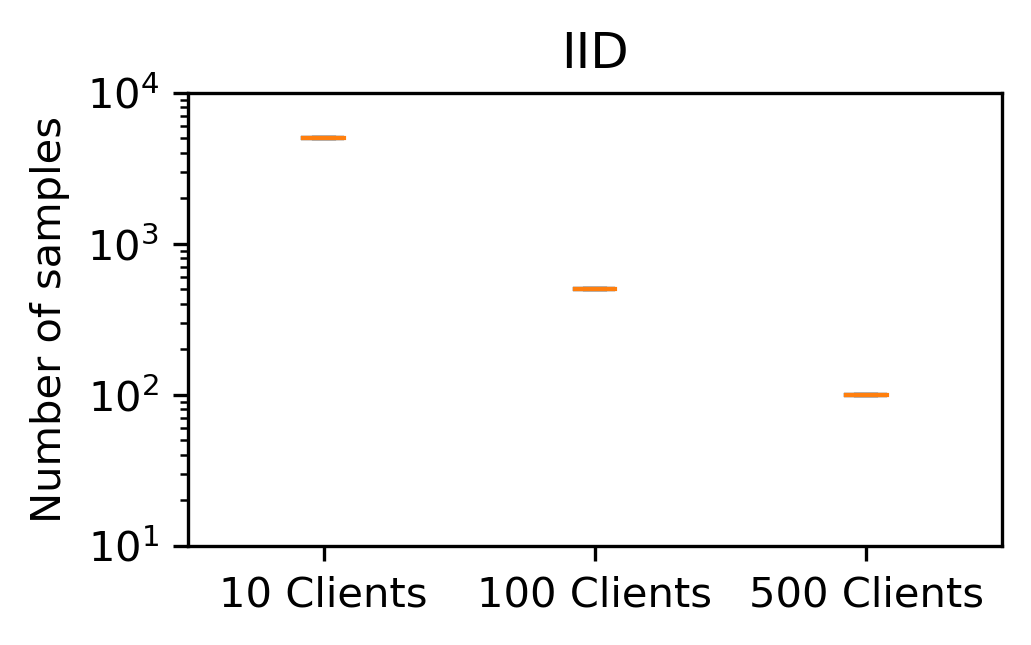}}
  \vspace{-10pt}
  \caption{\revise{The distribution of the number of samples per client $|\mathcal{D}_k|$ for CIFAR10.} }
  \label{fig:size}
  \vspace{-10pt}
\end{figure}

To better illustrate the data scarcity problem, we visualize the distribution for the local data size in Figure~\ref{fig:size}. As shown, when we distribute the training set of CIAFR10 ($50,000$ samples) to $10$ clients using a Dirichlet distribution parameterized by 0.1, these clients will present different class distributions, and the total number of local samples ranges from $2853$ to $8199$. This is the commonly explored non-IID setting. In this paper, we further explore scarce non-IID data, and thus we split the data into $100$ and $500$ clients. As a result,  the number of samples per client reduces significantly: the median number drops from $5685$ to $90$ when the number of clients increases from $10$ to $500$, as shown in Figure~\ref{fig:size}(a). This is the realistic scenario that we are interested in. It is also worth mentioning that data scarcity is independent of label skew and it can happen in the IID scenario. As shown in Figure~\ref{fig:size}(b), the local data covers $10$ classes uniformly, but the data scarcity problem becomes severe when the number of clients increases.

\subsection{Model Architecture and Hyper-parameters}\label{Sec:Apen_C2}

We classify images in CIFAR$10$ using MobileNet\_V2~\citep{sandler2018mobilenetv2}  that has $18$ blocks consisting of multiple convolutional and pooling layers.  The architectures of MobileNet\_V2 for CIFAR$10$ are summarized in Table~\ref{tab:mobi2}.

For audio classification, the audio samples are first transformed into spectrograms and fed into a CNN model, which we termed as \textit{AudioNet}. This model consists of four blocks, each comprising a convolutional layer, a Relu activation layer, and a Batch Normalization layer, followed by a fully connected layer\footnote{\url{https://www.kaggle.com/code/longx99/sound-classification/notebook}}. The details of the convolutional layers are summarized in Table~\ref{tab:audionet}.

For HAR data, we use a 4-layer CNN model, as summarized in Table~\ref{tab:harnet}. FLea shares the activation from the first layer. 

We use the Adam optimizer for local training with an initial learning rate of  $10^{-3}$  and decay it by $2\%$  per communication round until  $10^{-5}$. The size of the local batch is $32$, and we run $10$ local epochs for $100$ clients setting and $15$ local epochs for the rest.  For feature augmentation, we use $Beta(2,2)$. The weights in the loss function are set to $\lambda_1=1$ and $\lambda_2=3$. $10\%$ of clients are randomly sampled at each round. We run $100$ communications and take the best accuracy as the final result. For all results, we report the mean and standard deviation of the accuracy from five runs with different random seeds. 

\begin{table*}[t]
    \caption{Architecture of MobileNet\_V2. Features used to report the results in Table~\ref{table:result_all} are underlined.}
    \centering
    \resizebox{0.7\textwidth}{!}
    {%
    \begin{tabular}{c||cccccc}
    \toprule
        Block(CNN layers) &\#Input  &Operator & \#Output Channel &\#Kernel &\#Stride &\#Output \\\midrule
        $0(1)$ & $3\times32\times32$(image)   &conv2d &$32$ &$3$ &$1$ & $32\times32\times32$ \\
        $1(2-5)$ & $32\times32\times32$  &conv2d$\times4$ &$32,32,16,16$ &$1,3,1,1,$ &$1,1,1,1$ & \underline{$16\times32\times32$} \\
        $2(6-9)$ & $16\times32\times32$  &conv2d$\times4$ &$96,96,24,24$ &$1,3,1,1$ &$1,1,1,1$ & $ 32\times32\times32$\\
        $3(10-12)$ & $32\times32\times32$  &conv2d$\times3$ &$144,144,24$ &$1,3,1$ &$1,1,1$ & $24\times32\times32$\\
        $4(13-14)$ &  $24\times32\times32$   &conv2d$\times3$ &$144,144,32$ &$1,3,1$ &$1,2,1$ & $32\times16\times16$\\
        $5\&6(15-20)$ &  $32\times16\times16$   &conv2d$\times3$ &$192,192,32$ &$1,3,1$ &$1,1,1$ & $32\times16\times16$\\
        $7(21-23)$ &  $32\times16\times16$   &conv2d$\times3$ &$192,192,64$ &$1,3,1$ &$1,2,1$ & $64\times8\times8$\\
        $8,9,\&10(24-32)$ &  $64\times8\times8$   &conv2d$\times3$ &$384,384,64$ &$1,3,1$ &$1,1,1$ & $64\times8\times8$\\
        $11(33-36)$ &  $64\times8\times8$   &conv2d$\times4$ &$384,384,96,96$ &$1,3,,11$ &$1,1,1,1$ & $9\times8\times8$\\
        $12\&13(37-42)$ &  $96\times8\times8$   &conv2d$\times3$ &$576,576,96$ &$1,3,1$ &$1,1,1$ & $96\times8\times8$\\
        $14(43-45)$ &  $96\times8\times8$   &conv2d$\times3$ &$576,576,160$ &$1,3,1$ &$1,2,1$ & $160\times4\times4$\\
        $15\&16(46-51)$ &  $160\times4\times4$   &conv2d$\times3$ &$960,960,160$ &$1,3,1$ &$1,1,1$ & $160\times4\times4$\\
        $17(52-54)$ &  $160\times4\times4$   &conv2d$\times3$ &$960,960,320$ &$1,3,1$ &$1,1,1$ & $320\times4\times4$\\
        $18(55)$ & $320\times4\times4$   &conv2d &$1280$ &$1$ &$1$ & $1280\times4\times4$\\
    \bottomrule
    \end{tabular}
    }
    \label{tab:mobi2}
\end{table*}

\begin{table*}[t]
    \caption{\revise{Architecture of AudioNet. Features used to report the results in Table~\ref{table:result_all} are underlined.}}
    \centering
    \resizebox{0.7\textwidth}{!}
    {%
    \begin{tabular}{c||cccccc }
    \toprule
         Index &\#Input  &Operator & \#Output Channel &\#Kernel &\#Stride &\#Output\\\midrule
        $1$ & $2\times64\times344$(2-channel spectrogram)  &conv2d  &$8$ &$5$ &$2$ & $8\times32\times172$\\
        $2$ & $8\times32\times172$  &conv2d  &$16$ &$3$ &$2$ &  \underline{$16\times16\times86$}\\
        $3$ & $16\times16\times86$  &conv2d &$32$ &$3$ &$2$ & $32\times8\times43$\\
        $4$ &  $32\times8\times43$   &conv2d  &$64$ &$3$ &$2$ & $64\times4\times22$\\
        
    \bottomrule
    \end{tabular}
    }
    \label{tab:audionet}
\end{table*}

\begin{table*}[t]
    \caption{Architecture of decoder of MobileNet\_V2. }
    \centering
    \resizebox{0.7\textwidth}{!}
    {%
    \begin{tabular}{c||cccccc }
    \toprule
        Layer Index &\#Input  &Operator & \#Output Channel &\#Kernel &\#Stride &\#Output\\\midrule
        $1$ & $16\times32\times32$ (Feature)  &conv2d  &$32$ &$1$ &$1$ & $32\times32\times32$\\
        $2$ & $32\times32\times32$  &ConvTranspose2d  &$32$ &$3$ &$2$ &  $32\times64\times64$\\
        $3$ & $32\times64\times64$  &conv2d  &$32$ &$3$ &$2$ & $32\times32\times32$ \\
        $4$ & $32\times32\times32$  &conv2d &$3$ &$1$ &$1$ & $3\times32\times32$ (Data)\\   
    \bottomrule
    \end{tabular}
    }
    \label{tab:decoder}
\end{table*}

\begin{table*}[t]
    \caption{\revise{Architecture of HARNet. Features used to report the results in Table~\ref{table:result_all} are underlined.}}
    \centering
    \resizebox{0.5\textwidth}{!}
    {%
    \begin{tabular}{c||cccccc }
    \toprule
         Index &\#Input  &Operator & \#Output Channel &\#Kernel &\#Stride &\#Output\\\midrule
        $1$ & $1\times128\times3$  &conv2d  &$64$ &$3$ &$3$ & \underline{$64\times43\times5$}\\
        $2$ & $64\times43\times5$  &conv2d  &$128$ &$6$ &$2$ &  $128\times20\times7$\\
        $3$ & $16\times16\times86$  &conv2d &$256$ &$6$ &$2$ & $256\times9\times9$\\
        $4$ &  $256\times9\times9$   &conv2d  &$256$ &$6$ &$2$ & $256\times3\times11$\\
        
    \bottomrule
    \end{tabular}
    }
    \label{tab:harnet}
\end{table*}

\subsection{Baseline Implementation}\label{Sec:Apen_C3}

More details for baseline implementations are summarized as blew,
\begin{itemize}[topsep=0pt, itemsep=0pt,  leftmargin=15pt]
\item \textbf{\emph{FedProx}}: We adapt the implementation from~\citep{li2020federated}. We test the weight for local model regularization in $[0.1,0.01,0.001]$  and report the best results.  
\item \textbf{\emph{FedLC}}: it calibrates the logits before softmax cross-entropy according to the probability of occurrence of each class~\citep{zhang2022federated}. We test the scaling factor in the calibration from 0.1 to 1 and report the best performance.  
\item \textbf{FedDecorr}: This method applies a regularization term during local training that encourages different dimensions of the low-dimensional features to be uncorrelated~\citep{shi2022towards}.  We adapted the official implementation\footnote{\url{https://github.com/bytedance/FedDecorr}} and suggested hyper-parameter in the source paper. We found that this method can only outperform \textit{FedAvg} with fewer than $10$ clients for CIFAR$10$. 
\item \textbf{\emph{FedNTD}}: It prevents the local model drift by distilling knowledge from the global model~\citep{lee2022preservation}. We use the default distilling weights from the original paper as the settings are similar\footnote{\url{https://github.com/Lee-Gihun/FedNTD.git}}. 
\item\textbf{FedBR}~\citep{guo2023fedbr}: this approach leverage $32$ mini-batch data averages without class labels as data augmentation. A min-max algorithm is designed, where the max step aims to make local features for all classes more distinguishable. In contrast, the min step enforces the local classifier to predict uniform probabilities for the global data averages. We adopt the official implementation\footnote{\url{https://github.com/lins-lab/fedbr}} in our framework. 
\item \textbf{\emph{CCVR}}: It collects a global feature set before the final fully connected linear of the converged global model, i.e., the model trained via \emph{FedAvg}, to calibrate the classifier on the server~\citep{luo2021no}. For a fair comparison, we use the same amount of features as our method for this baseline, and we fit the model using the features instead of distributions as used in~\citep{luo2021no}.  This allows us to report the optimal performance of \emph{CCVR}. 
\item \textbf{FedGen}: It is a method that trains a data generator using the global model as the discriminator to create synthetic data for local training~\citep{liu2022overcoming}. The generator outputs $\hat{x}_i$ with input $(y_i,z_i)$ where $z_i$ is a sample for Normal distribution. The generator is a convolutional neural network consisting of four \emph{ConvTranspose2d} layers to upsample feature maps. We train the first 30 rounds by normal \emph{FedAvg} and after 30 rounds, we use the global model as 
the discriminator to distinguish with the generated data $\hat{x}_i$ is real or not.  
\item \textbf{FedData}: In this baseline, we assume the server waits until all the clients have shared $10\%$ of their local data in the beginning round. The gathered data will be sent to clients to mix with local data for model training. 
\item \textbf{FedMix}: Similar to \textit{FedData}, we assume the server waits until all the clients have shared their data averages. we use a mini-batch of $10$ to aggregate the samples. Different from \textit{FedBR} The gathered data will be sent to clients, combined with local data based on the Beta distribution.
\end{itemize}

\section{Privacy study}\label{sec:attacks}
 
Now we present the experimental setup for privacy attacks. We use the Qua($3$) data splits when $|K|=100$ as an example for studying, as in other settings either the label is more skewed or the local data is more scarce,  privacy attack can hardly be more effective than this setting. This is to present the attack defending for the most vulnerable case. As the correlation between the features and the data is continuously reduced (shown in Figure~\ref{fig:corr}), we report the reconstruction and context detection performance for $c=0.65$ (the $1^{st}$ round) and $c=0.40$ (the $10^{th}$ round) for reference.


\textit{\textbf{Data reconstruction}}. We first implemented a data reconstruction attacker, following the approach described in~\cite{dosovitskiy2016inverting}, the attacker constructed a decoder for reconstruction. Specifically, the attacker targeted the converged global model, ensuring a fair comparison. The decoder architecture, designed to match the MobileNet\_V2 architecture, comprised four \emph{conv2d} layers (refer to Table~\ref{tab:decoder}) to reconstruct the original data from the provided features. For visualization purposes, the CIFAR$10$ images were cropped to a size of $32\times32$ pixels without any normalization. The decoder took the features extracted from the global model as input and generated a reconstructed image, which served as the basis for calculating the mean squared error (MSE).

To train the decoder, we utilized the entire CIFAR$10$ training set, conducting training for $20$ epochs and employing a learning rate of $0.001$. This approach allowed us to evaluate the fidelity of the reconstructed data and compare it with the original input, providing insights into the effectiveness of our proposed feature interpolation method.  We use the testing set and the target global model ($c=0.65$  and $c=0.40$ ) to extract features for reconstruction. Figure~\ref{fig:corr} shows the training MSE while the exampled images are from the testing set. For $c=0.65$, i.e., after the first round, the sensitive attributes are removed (e.g., the color of the dog). After $10$ rounds when $c<0.4$, information is further compressed and the privacy protection is enhanced. 
Overall, with $\mathcal{L}_{dec}$, the correlation between data and features is reduced, preventing the image from being reconstructed.

\textit{\textbf{Identifying context information}}.
In this attack, we assume that the attacker explicitly knows the context information and thus can generate large amounts of negative (clear data) and positive (clear data with context marker) pairs to train a context classifier (which is very challenging and unrealistic but this is for the sake of testing).  Real-world attacks will be far more challenging than our simulations.

The context identification attacker is interested in finding out if a given feature  $f$,  is from the source data with a specific context or not. 
We simulate the context information by adding a color square to the image (to mimic the camera broken), as illustrated in Figure~\ref{fig:datashared}.  We use a binary classifier consisting of four linear layers to classify the flattened features or images. 
 To train the classifier, we add the context marker to half of the training set. To report the identification performance, we add the same marker to half of the testing set. In Figure~\ref{fig:recons}, the identification accuracy for \textit{FedMix} and our \textit{FLea} are given. We measure the attacking difficulty by how many training samples the model needs to achieve a certain accuracy. The results in Figure~\ref{fig:recons} suggest that \textit{FLea} needs times of training sample than \textit{FedMix} for different correlations. This demonstrates that \textit{FLea} can better protect the context privacy.



All the above results lead to the conclusion that by reducing feature exposure and mitigating the correlation between the features and source data,  \textit{FLea} safely protects the privacy associated with feature sharing while achieving favorable performance gain in addressing the label skew and data scarcity simultaneously. 


\end{document}